\documentclass{article}
\usepackage{conference,times}

\usepackage{graphicx}
\usepackage{algorithm}
\usepackage{algorithmic}
\usepackage{amsmath,amsfonts,bm}

\def\eqref#1{equation~\ref{#1}}

\def\1{\bm{1}}

\DeclareMathAlphabet{\mathsfit}{\encodingdefault}{\sfdefault}{m}{sl}
\SetMathAlphabet{\mathsfit}{bold}{\encodingdefault}{\sfdefault}{bx}{n}

\usepackage{hyperref}
\usepackage{url}
\usepackage{booktabs,multirow,subcaption}
\usepackage{siunitx}
\usepackage{float}
\usepackage{placeins}
\usepackage{tikz}
\usepackage{fontawesome5}

\newcommand{\projecturl}{https://akshaymete.com/projects/optta}
\newcommand{\codeurl}{https://github.com/akshaymete/optta}
\newcommand{\linkbutton}[3]{{\hypersetup{pdfborder={0 0 0}}\href{#3}{\tikz[baseline=(b.base)]\node[draw=black!20, fill=black!4, rounded corners=4pt, inner xsep=5pt, inner ysep=2.5pt, text=black!75, font=\sffamily\small](b){#1\enspace#2};}}}

\title{Adapt for Planning, Not Just Prediction:\\ Optimistic Test-Time World Model Adaptation}

\author{Akshay Mete, Dileep Kalathil, P. R. Kumar \\
ECE Department, Texas A\&M University, College Station, Texas, USA. \\
\texttt{\{akshaymete,dileep.kalathil,prk\}@tamu.edu }
}

\begin{document}
\addtocontents{toc}{\protect\setcounter{tocdepth}{-1}}

\maketitle

\vspace{-2em}
\begin{center}
\linkbutton{\faGlobe}{Webpage}{\projecturl}\hspace{1em}\linkbutton{\faGithub}{Code}{\codeurl}
\end{center}

\begin{abstract}
Latent world models are typically pre-trained offline and kept frozen at test time, so planning with them can degrade under distribution shift. Test-time adaptation (TTA) can mitigate this degradation by updating the model during deployment. Unlike pre-training on fixed data, TTA for planning is closed-loop: the planner uses model rollouts to select actions, and the resulting transitions drive the next update. This raises the question: which objective should guide TTA for planning? Under a prediction-only objective, the model is fit only to the outcomes of its own actions, so it can remain wrong about the actions that would reach the goal, and low prediction error need not translate into better planning. Inspired by reward-biased maximum likelihood from adaptive control, we introduce Optimistic Test-Time Adaptation (OpTTA). Its objective has two terms, the prediction loss on observed transitions and the planner's goal-reaching cost on its candidate rollouts, both minimized over the model parameters. The planning term biases adaptation toward models that are optimistic about reaching the goal, without any additional environment interaction. Across visual and environment shifts on manipulation, reaching and navigation tasks, OpTTA raises final planning success over the frozen model by 13.5 percentage points on average, compared with 6.9 for the prediction-only objective (AdaJEPA), when only the predictor is adapted. When the encoder is adapted as well, the gains are 13.5 and 7.5 points. These experiments span three JEPA world models with different encoders and training objectives and, for temporal straightening, both CEM and gradient-based planners. Our results suggest that test-time adaptation of world models for planning benefits from an optimistic, task-aware term in addition to the prediction loss.
\end{abstract}

\section{Introduction}

\begin{figure}[!b]
  \centering
  \includegraphics[width=\textwidth]{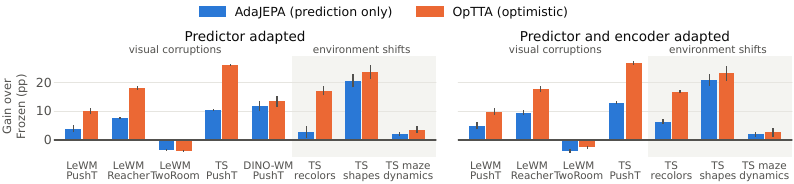}
  \caption{Gain in final planning success over the frozen model for prediction-only adaptation (AdaJEPA) and optimistic adaptation (OpTTA), averaged over the visual corruptions or environment shifts of each setting (Table~\ref{tab:main}). Error bars show $\pm1$ s.e.\ over seeds.}
  \label{fig:gains}
\end{figure}

\begin{figure}[t]
  \centering
  \includegraphics[width=\textwidth]{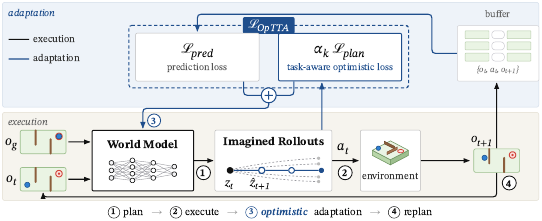}
  \caption{Optimistic Test-Time Adaptation. At every replanning step the world model is adapted on the observed transitions, using the prediction loss plus the task-aware optimistic loss evaluated on the candidate rollouts the planner has already generated.}
  \label{fig:main_loop}
\end{figure}

Latent world models have demonstrated success across a range of control and planning tasks \citep{lecun2022path,zhou2025dino,maes2026leworldmodel}.
These world models are typically pre-trained offline and kept frozen at test time.
Test-time distribution shift can degrade planning performance.
A standard approach in adaptive control is to update the dynamics model online as new transitions are observed.
This raises the central question: which objective should guide test-time adaptation of a world model?

A straightforward choice is to minimize the latent prediction loss on transitions observed at test time.
AdaJEPA \citep{wang2026adajepa} follows this approach for goal-conditioned planning with JEPA world models, updating the model on recent transitions at each replanning step.
Compared with a frozen model, this prediction-only adaptation can improve planning success under distribution shift.
Unlike pre-training on fixed data, test-time adaptation operates in a closed loop: the planner selects actions by scoring the model's imagined rollouts, and the resulting transitions are used for the next model update. The model is thus trained to minimize prediction loss only on the outcomes of its own action choices, so it can fit them accurately yet remain wrong about actions that would reach the goal, a closed-loop identifiability problem \citep{borkar1979adaptive}.
This mismatch between the prediction loss and the planning objective is known as objective mismatch \citep{pmlr-v120-lambert20a}.

An early solution to this objective mismatch was reward-biased maximum likelihood (RBMLE), proposed by \citet{kumar1982rbmle}.
RBMLE estimates the model with a joint objective of two terms: the likelihood of the observed transitions and the optimal long-run reward achievable under the model.
This objective is both task-aware and optimistic: it favors a model with high achievable reward while anchoring the model estimate to the observed transitions.
RBMLE introduces optimism directly through the estimation objective, without the confidence sets or explicit uncertainty estimates required by upper-confidence-bound methods \citep{jaksch2010ucrl}, making it more amenable to test-time adaptation, where calibrated uncertainty estimates are difficult to obtain for world models under distribution shift.
RBMLE has so far been studied mainly in adaptive control and reinforcement learning \citep{kumar1982rbmle,mete2021rbmle,mete2026optimistic}; in this paper, we propose an optimistic objective for test-time adaptation of world models used for planning.

\subsection*{Contributions}

\paragraph{Optimistic Test-Time Adaptation (OpTTA).}
We propose the following objective for model adaptation, which balances prediction accuracy on observed transitions against the goal-reaching cost of the model's own imagined rollouts:
\begin{equation*}
    \mathcal{L}_{\rm OpTTA} = \mathcal{L}_{\rm pred} + \alpha_k\,\mathcal{L}_{\rm plan}, \qquad \alpha_k \geq 0.
\end{equation*}
The prediction term, $\mathcal{L}_{\rm pred}$, is the model's latent prediction loss evaluated on recently observed test-time transitions. The planning term, $\mathcal{L}_{\rm plan}$, is the planner's goal-reaching cost evaluated on its imagined rollouts, with gradients taken with respect to the world-model parameters rather than the actions. Here $\alpha_k$ is the optimism weight at replanning step $k$. Choosing $\alpha_k>0$ makes adaptation optimistic by favoring model parameters under which the imagined rollouts have lower goal-reaching cost, while $\mathcal{L}_{\rm pred}$ anchors the model to the observed transitions; setting $\alpha_k=0$ recovers AdaJEPA's prediction-only objective.

\paragraph{Optimism improves planning under distribution shift.}
Across three JEPA world models (LeWM, DINO-WM and temporal straightening) and 15 shifted settings spanning visual corruptions, recolorings, held-out shapes, and changes in mass and damping, optimistic adaptation (OpTTA) raises final success over the frozen model by $13.5$ points on average, compared with $6.9$ for prediction-only adaptation (AdaJEPA); with the encoder also adapted, the gains are $13.5$ and $7.5$ (Table~\ref{tab:main}; Figure~\ref{fig:gains}).

\section{Related Work}

\paragraph{Test-time adaptation of world models.}
Test-time adaptation is used in a wide variety of settings to address distribution shift \citep{sun2020test,wang2020tent,hansen2020self,liang2025comprehensive}.
Several recent works adapt world models to distribution shift \citep{gao2025adaworld,wang2025adawm,wan2026worldagen,park2026test,wang2026adajepa}.
For latent world models used in planning, AdaJEPA \citep{wang2026adajepa} proposed test-time adaptation by minimizing prediction error on observed transitions.
Updating the dynamics model online as new transitions are observed is also standard in adaptive control and model-based reinforcement learning \citep{borkar1979adaptive,kumar1982rbmle,hafner2020dream,hafner2024mastering,hansen2024td}.

\paragraph{Objective mismatch.}
A large number of latent world models are trained and adapted to minimize prediction error, but are then used to select actions for planning or control \citep{hafner2020dream,hafner2024mastering,zhou2025dino,wang2026temporal,maes2026leworldmodel,wang2026adajepa}.
This leads to a mismatch between the model-learning objective and the control objective \citep{pmlr-v120-lambert20a}.
To address this mismatch, value-aware model-learning methods train the dynamics model using reward or value information in addition to prediction error \citep{farahmand2017vaml,farahmand2018itervaml,hansen2024td}.
These methods require a learned reward or value function, which is unavailable when adapting a pretrained world model for goal-conditioned planning.
In this paper, we argue that optimism offers a practical solution to objective mismatch in test-time world-model adaptation for goal-conditioned planning.

\paragraph{Reward-biased maximum likelihood estimation.}
Reward-biased maximum likelihood estimation (RBMLE) \citep{kumar1982rbmle} was proposed to counteract the pessimistic bias that maximum-likelihood estimation can exhibit in closed-loop control \citep{borkar1979adaptive}.
It favors models with higher optimal reward while anchoring the estimate to the observed transitions.
Later, upper-confidence-bound (UCB) methods implemented optimism by optimizing over a confidence set of plausible models \citep{auer2002finite,jaksch2010ucrl}, of which RBMLE can be viewed as a Lagrangian relaxation \citep{mete2022augmented}.
Uncertainty-based optimistic approaches \citep{curi2020efficient,sukhija2025sombrl} rely on calibrated uncertainty estimates, which are difficult to obtain under distribution shift and limited deployment data, whereas RBMLE requires none.
RBMLE has since been extended to reinforcement learning \citep{mete2021rbmle}, linear-quadratic control \citep{mete2022augmented}, and, recently, online world-model-based reinforcement learning \citep{mete2026optimistic}.
Whereas \citet{mete2026optimistic} apply RBMLE-style optimism to online reinforcement learning with rewards, OpTTA applies it to test-time adaptation of pretrained world models for goal-conditioned planning.

\paragraph{JEPA world models.}
Joint-embedding predictive architectures learn representations by predicting target embeddings rather than reconstructing observations \citep{lecun2022path,assran2023self,assran2025v}.
JEPA world models such as PLDM, DINO-WM, temporal straightening, and LeWM \citep{sobal2026learning,zhou2025dino,wang2026temporal,maes2026leworldmodel} apply this principle to action-conditioned prediction for goal-conditioned planning.
In this paper, we address test-time adaptation of these world models for goal-conditioned planning.

\section{Preliminaries}
\paragraph{Latent world models.}
We consider a pretrained latent world model comprising an observation encoder $\mathcal{E}^s_\phi$, an action embedding $\mathcal{E}^a_\psi$, and a latent dynamics predictor $f_\theta$:
\begin{equation*}
    z_t = \mathcal{E}^s_\phi(o_t), \qquad u_t = \mathcal{E}^a_\psi(a_t), \qquad \hat z_{t+1} = f_\theta(z_t, u_t).
\end{equation*}
Here, $o_t$ and $a_t$ denote the observation and action at time $t$, while $z_t$ and $u_t$ denote the corresponding latent state and action embedding.
For compactness, we write $f_\theta(z_t,u_t)$ while suppressing its dependence on the $L$ most recent latent states and action embeddings.
Rolling out the predictor from latent state $z_t$ under an action sequence $a_{t:t+n-1}$ of length $n$ yields the imagined latent-state trajectory
\begin{equation}
    \hat z_{t+k} = f_\theta(\hat z_{t+k-1}, u_{t+k-1}), \quad k = 1,\dots,n, \qquad \hat z_t = z_t.
    \label{eq:rollout}
\end{equation}
We denote this trajectory by $\hat z_{t+1:t+n} = \tau_\theta(z_t, a_{t:t+n-1})$; it is differentiable with respect to both the action sequence and the predictor parameters $\theta$.

\paragraph{Goal-conditioned planning.}
At test time, the task is given by a goal image $o_g$, encoded as $z_g = \mathcal{E}^s_\phi(o_g)$.
Under model predictive control (MPC) with horizon $H$, the agent scores each candidate action sequence by the weighted goal-distance cost of its imagined latent-state trajectory:
\begin{equation}
    J(a_{t:t+H-1},\, \theta) = \sum_{k=1}^{H} w_k\, d(\hat z_{t+k},\, z_g), \qquad \hat z_{t+1:t+H} = \tau_\theta(z_t, a_{t:t+H-1}),
    \label{eq:goal_cost}
\end{equation}
where the weights $w_k \ge 0$ set the relative importance of each step and $d$ is the squared Euclidean distance.
At each replanning step, let $\{a^{(m)}_{t:t+H-1}\}_{m=1}^{M}$ denote the candidate action sequences retained for adaptation. We collect each sequence and its induced latent-state trajectory in the rollout population
\begin{equation}
    \mathcal{P}_\theta
    =
    \left\{
    \left(
    a^{(m)}_{t:t+H-1},
    \tau_\theta\!\left(z_t,a^{(m)}_{t:t+H-1}\right)
    \right)
    \right\}_{m=1}^{M}.
    \label{eq:rollout_population}
\end{equation}
During adaptation, the action sequences are held fixed while their latent-state trajectories are recomputed as $\theta$ changes.
Planning is performed using either the cross-entropy method (CEM) \citep{rubinstein1997optimization} or gradient descent.

\paragraph{Test-time adaptation.}
Under test-time distribution shift, a frozen pretrained world model can produce inaccurate rollouts, degrading planning performance.
AdaJEPA \citep{wang2026adajepa} performs test-time adaptation for goal-conditioned planning by updating the world model from a buffer $\mathcal{B}$ containing up to the $N$ most recent observed transitions $(o_i,a_i,o_{i+1})$.
It minimizes the one-step latent prediction loss
\begin{equation}
    \mathcal{L}_{\rm pred}(\mathcal{B}) = \frac{1}{|\mathcal{B}|} \sum_{(o_i, a_i, o_{i+1}) \in \mathcal{B}} \ell\big(f_\theta(z_i, u_i),\ \operatorname{sg}(z_{i+1})\big).
    \label{eq:tta_loss}
\end{equation}
Here, $\ell$ is the mean-squared error in latent space.
A designated set $\Omega$ of observation-encoder and/or predictor parameters is updated for $U$ gradient steps:
\begin{equation}
    \Omega \leftarrow \Omega - \eta \nabla_\Omega \mathcal{L}_{\rm pred}(\mathcal{B}).
    \label{eq:tta_update}
\end{equation}
The updated model is used for the next replanning step.
Together, these steps form a plan--execute--adapt--replan loop.
Adaptation is task-local: after each task, the model is reset to its pretrained parameters and the buffer is cleared.

\section{Optimistic Test-Time Adaptation for Latent World Models}
\label{sec:method}

Inspired by RBMLE, we formulate test-time adaptation of latent world models as task-aware, optimistic model estimation.
At replanning step $k$, the OpTTA model-estimation objective has two components: prediction error on the observed transitions $\mathcal{B}$ and goal-reaching cost on the candidate rollouts $\mathcal{P}_\theta$:
\begin{equation}
    \mathcal{L}_{\rm OpTTA}(\mathcal{B},\mathcal{P}_\theta)
    = \mathcal{L}_{\rm pred}(\mathcal{B})
    + \alpha_k\,\mathcal{L}_{\rm plan}(\mathcal{P}_\theta).
    \label{eq:optta_objective}
\end{equation}
Here, $\mathcal{L}_{\rm pred}$ fits the model to observed transitions, whereas $\mathcal{L}_{\rm plan}$ evaluates the goal-reaching cost of imagined rollouts under candidate action sequences.
The planning component is the average goal-reaching cost of the candidate rollouts:
\begin{equation}
    \mathcal{L}_{\rm plan}(\mathcal{P}_\theta)
    = \frac{1}{M}\sum_{m=1}^{M}
    J\!\left(a^{(m)}_{t:t+H-1},\theta\right).
    \label{eq:planning_loss}
\end{equation}
The coefficient $\alpha_k\ge 0$ controls the strength of optimism at replanning step $k$.

With $\alpha_k>0$, $\mathcal{L}_{\rm plan}$ favors model estimates that assign lower goal-reaching costs to candidate rollouts, while $\mathcal{L}_{\rm pred}$ keeps those estimates grounded in observed transitions.
The objective does not assume a particular latent world-model architecture or planner.
We instantiate OpTTA within the AdaJEPA protocol, retaining its planner, transition buffer, adapted parameter set, and update schedule, but replacing its prediction-only adaptation objective with \eqref{eq:optta_objective}.

Thus, OpTTA leaves three configurable design choices: which model parameters to update, which imagined rollouts to include in $\mathcal{P}_\theta$, and how to set $\alpha_k$.

\paragraph{Which model parameters are updated?}
Following AdaJEPA, OpTTA adapts a selected parameter set $\Omega$ comprising parameters of the observation encoder, the latent predictor, or both, while keeping the action-embedding parameters fixed.
$\mathcal{L}_{\rm pred}$ is differentiated through all parameters in $\Omega$, whereas $\mathcal{L}_{\rm plan}$ is differentiated only through the predictor parameters in a selected subset $\Omega_f\subseteq\Omega$.

\paragraph{Which imagined rollouts form $\mathcal{P}_\theta$?}
In our instantiation, $\mathcal{P}_\theta$ contains the $M$ candidate action sequences of CEM's final iteration at the current planning step, each with its imagined latent-state trajectory.
During adaptation, the candidate action sequences are held fixed, and $\mathcal{L}_{\rm plan}$ is differentiated through their model-predicted trajectories with respect to $\Omega_f$, rather than the actions.

\paragraph{How is $\alpha_k$ chosen?}
The gradient magnitudes of $\mathcal{L}_{\rm pred}$ and $\mathcal{L}_{\rm plan}$ can vary substantially across world models, planners, tasks, distribution shifts, and replanning steps, so a fixed optimism weight can cause either term to dominate.
We therefore set $\alpha_k$ so that, on $\Omega_f$, the weighted planning gradient has approximately a scheduled magnitude relative to the prediction gradient:
\begin{equation}
    \alpha_k
    = \rho\sqrt{\frac{k}{K_{\max}}}
    \frac{
        \left\lVert \nabla_{\Omega_f}\mathcal{L}_{\rm pred}(\mathcal{B}) \right\rVert_2
    }{
        \left\lVert \nabla_{\Omega_f}\mathcal{L}_{\rm plan}(\mathcal{P}_\theta) \right\rVert_2
        + \varepsilon
    }.
    \label{eq:alpha}
\end{equation}
Here, $\rho\ge 0$ controls the relative gradient contribution of the planning term, while $\sqrt{k/K_{\max}}$ limits optimism early and increases it over replanning steps.
The stabilizer $\varepsilon>0$ prevents division by zero.
We treat $\alpha_k$ as a constant when differentiating $\mathcal{L}_{\rm OpTTA}$ with respect to the model parameters.
We analyze how each of these design choices affects planning performance in Section~\ref{sec:ablations}.
With these design choices specified, Algorithm~\ref{alg:optta} summarizes the resulting OpTTA procedure.

\begin{algorithm}[t]
\caption{Optimistic Test-Time Adaptation for one task}
\label{alg:optta}
\small
\begin{algorithmic}[1]
\STATE \textbf{Input:} pretrained model $(\mathcal{E}^s_\phi,\mathcal{E}^a_\psi,f_\theta)$, planner, goal observation $o_g$, adapted parameters $\Omega$, and planning-loss parameters $\Omega_f\subseteq\Omega\cap\theta$
\STATE \textbf{Hyperparameters:} base optimism $\rho$, stabilizer $\varepsilon$, replanning budget $K_{\max}$, adaptation steps $U$, buffer capacity $N$, and learning rate $\eta$
\STATE Restore the pretrained parameters; $\mathcal{B}\leftarrow\emptyset$; $t\leftarrow 0$; observe $o_0$
\FOR{$k=0,\ldots,K_{\max}-1$}
    \STATE Encode the current observation and the goal: $z_t\leftarrow\mathcal{E}^s_\phi(o_t)$, $z_g\leftarrow\mathcal{E}^s_\phi(o_g)$
    \STATE $(a^\star_{t:t+H-1},\mathcal{P})\leftarrow\operatorname{Plan}(z_t,z_g;\mathcal{E}^a_\psi,f_\theta)$ \hfill \COMMENT{$\mathcal{P}$: candidates of CEM's final iteration}
    \STATE Execute $a^\star_t$, observe $o_{t+1}$, and append $(o_t,a^\star_t,o_{t+1})$ to $\mathcal{B}$, retaining the $N$ most recent transitions
    \IF{the goal is reached}
        \STATE \textbf{stop}
    \ENDIF
    \FOR{$u=1,\ldots,U$}
        \STATE Compute $\mathcal{L}_{\rm pred}(\mathcal{B})$ and $\mathcal{L}_{\rm plan}(\mathcal{P})$
        \STATE Compute $\alpha_k$ using \eqref{eq:alpha} and stop its gradient
        \STATE $\Omega\leftarrow\Omega-\eta\nabla_\Omega\!\bigl[\mathcal{L}_{\rm pred}+\alpha_k\mathcal{L}_{\rm plan}\bigr]$ \hfill \COMMENT{$\mathcal{L}_{\rm plan}$ gradients are restricted to $\Omega_f$}
    \ENDFOR
    \STATE $t\leftarrow t+1$
\ENDFOR
\end{algorithmic}
\end{algorithm}

\section{Experiments}
\label{sec:experiments}

Our experiments evaluate the benefits of optimistic, task-aware world-model adaptation for goal-conditioned planning under distribution shift.
The evaluation spans three latent world models; manipulation, reaching, and navigation tasks; visual and environment shifts; predictor-only and joint predictor--encoder adaptation; and sampling-based and gradient-based planners.

\subsection{Setup}
\label{sec:setup}
\paragraph{World models and tasks.}
We evaluate three latent world models: LeWM, temporal straightening, and DINO-WM.
We evaluate goal-conditioned planning on tasks commonly used in related world-model work \citep{zhou2025dino,wang2026temporal,maes2026leworldmodel}: object manipulation (PushT and PushObj), reaching (Reacher), and navigation (TwoRoom and maze tasks).

\paragraph{Distribution shifts.}
We evaluate seven visual corruptions (blur, salt-and-pepper noise, darkening, gamma, hue, contrast, and translation) and environment shifts in object color, geometry, mass, and damping.
Severities are selected per checkpoint using only the frozen model.
LeWM and temporal straightening results average over four severities and five seeds per corruption; DINO-WM uses one severity and three seeds.

\paragraph{Controlled comparison.}
We compare prediction-only TTA (AdaJEPA \citep{wang2026adajepa}) and OpTTA against the frozen pretrained model.
Both TTA methods share the initialization, planner, transition buffer, adapted parameters, and update schedule, so they differ only in the adaptation objective.
We adapt either the predictor alone or the predictor and encoder jointly; DINO-WM, whose DINOv2 encoder is frozen, uses predictor adaptation only.

\paragraph{Protocol and metrics.}
Unless noted, CEM plans five actions, executes one, and replans at most 30 times.
All methods share 50 start--goal episodes per seed.
We report success at replan 30 (\emph{Final}) and its average over replans 1--30 (\emph{Mean over replans}), which measures how early tasks are solved.
OpTTA uses \eqref{eq:alpha} with $\rho=1$ throughout, without setting-specific tuning.

\subsection{Planning performance}
\label{sec:main}
\paragraph{Under distribution shift.}
Across both adaptation scopes, OpTTA achieves higher final success than AdaJEPA in 14 of 15 shifted settings (Table~\ref{tab:main}).
Across the eight predictor-only settings, OpTTA improves over Frozen by $13.5$ points on average, versus $6.9$ for AdaJEPA; across the seven joint predictor--encoder settings, the gains are $13.5$ and $7.5$ points.

\providecommand{\gain}[1]{\,{\scriptsize(#1)}}

\providecommand{\gain}[1]{\,{\scriptsize(#1)}}
\begin{table}[t]
\centering
\small
\setlength{\tabcolsep}{2.8pt}
\renewcommand{\arraystretch}{0.96}
\setlength{\aboverulesep}{0.3ex}
\setlength{\belowrulesep}{0.45ex}
\caption{Planning success (\%, mean $\pm$ s.e.\ over seeds). Final is success by replanning step $30$, and Mean over replans is the average cumulative success rate across replanning steps $1$--$30$. Parentheses report the adaptation gain over Frozen, paired by seed and computed before rounding. Panel (a) adapts only the predictor; panel (b) adapts both the predictor and encoder. Shifted rows average over the number of shifts in their labels, each visual corruption over four severity levels (DINO-WM: one level and three seeds); per-shift results are in Appendix~\ref{app:experiments}.}
\label{tab:main}
\begin{subtable}{\textwidth}
\centering
\caption{Predictor adapted.}
\label{tab:main_pred}
\begin{tabular}{l ccc @{\hspace{9pt}} ccc}
\toprule
 & \multicolumn{3}{c}{Final} & \multicolumn{3}{c}{Mean over replans} \\
\cmidrule(lr){2-4}\cmidrule(lr){5-7}
Setting (shifts) & Frozen & AdaJEPA & OpTTA & Frozen & AdaJEPA & OpTTA \\
\midrule
\multicolumn{7}{l}{\textit{In distribution}} \\
LeWM PushT            & 80.0{\scriptsize$\pm$2.5} & 88.8{\scriptsize$\pm$1.4}\gain{+8.8} & \textbf{92.0}{\scriptsize$\pm$1.9}\gain{+12.0} & 59.1{\scriptsize$\pm$1.1} & 66.7{\scriptsize$\pm$1.3}\gain{+7.6} & \textbf{71.5}{\scriptsize$\pm$1.9}\gain{+12.4} \\
LeWM Reacher          & 98.4{\scriptsize$\pm$0.7} & 98.8{\scriptsize$\pm$0.8}\gain{+0.4} & \textbf{100.0}{\scriptsize$\pm$0.0}\gain{+1.6} & 75.6{\scriptsize$\pm$0.9} & 75.3{\scriptsize$\pm$1.5}\gain{$-$0.3} & \textbf{77.6}{\scriptsize$\pm$0.6}\gain{+2.0} \\
LeWM TwoRoom          & \textbf{93.6}{\scriptsize$\pm$1.3} & 93.2{\scriptsize$\pm$1.2}\gain{$-$0.4} & 91.2{\scriptsize$\pm$0.5}\gain{$-$2.4} & \textbf{80.7}{\scriptsize$\pm$0.8} & 79.7{\scriptsize$\pm$1.0}\gain{$-$1.0} & 79.0{\scriptsize$\pm$0.7}\gain{$-$1.7} \\
TS PushT              & 68.4{\scriptsize$\pm$2.3} & 81.6{\scriptsize$\pm$2.5}\gain{+13.2} & \textbf{91.6}{\scriptsize$\pm$1.9}\gain{+23.2} & 49.4{\scriptsize$\pm$1.2} & 55.9{\scriptsize$\pm$1.1}\gain{+6.5} & \textbf{66.1}{\scriptsize$\pm$1.3}\gain{+16.7} \\
DINO-WM PushT         & 76.7{\scriptsize$\pm$3.5} & 77.3{\scriptsize$\pm$5.2}\gain{+0.7} & \textbf{81.3}{\scriptsize$\pm$4.7}\gain{+4.7} & 58.9{\scriptsize$\pm$2.2} & 59.7{\scriptsize$\pm$5.0}\gain{+0.8} & \textbf{63.1}{\scriptsize$\pm$3.9}\gain{+4.2} \\
\midrule
\multicolumn{7}{l}{\textit{Visual corruptions}} \\
LeWM PushT (7)        & 58.5{\scriptsize$\pm$0.8} & 62.4{\scriptsize$\pm$1.0}\gain{+3.9} & \textbf{68.5}{\scriptsize$\pm$0.6}\gain{+10.0} & 43.9{\scriptsize$\pm$0.7} & 47.3{\scriptsize$\pm$0.6}\gain{+3.5} & \textbf{52.9}{\scriptsize$\pm$0.5}\gain{+9.0} \\
LeWM Reacher (7)      & 52.8{\scriptsize$\pm$0.5} & 60.3{\scriptsize$\pm$0.5}\gain{+7.5} & \textbf{70.8}{\scriptsize$\pm$0.7}\gain{+18.0} & 38.4{\scriptsize$\pm$0.6} & 40.6{\scriptsize$\pm$0.5}\gain{+2.2} & \textbf{47.2}{\scriptsize$\pm$0.8}\gain{+8.8} \\
LeWM TwoRoom (7)      & \textbf{61.0}{\scriptsize$\pm$1.7} & 57.6{\scriptsize$\pm$1.5}\gain{$-$3.4} & 57.2{\scriptsize$\pm$1.5}\gain{$-$3.8} & \textbf{51.9}{\scriptsize$\pm$1.8} & 49.3{\scriptsize$\pm$1.6}\gain{$-$2.6} & 49.3{\scriptsize$\pm$1.5}\gain{$-$2.6} \\
TS PushT (7)          & 47.8{\scriptsize$\pm$1.5} & 58.3{\scriptsize$\pm$1.2}\gain{+10.4} & \textbf{74.0}{\scriptsize$\pm$1.3}\gain{+26.1} & 34.0{\scriptsize$\pm$1.3} & 38.2{\scriptsize$\pm$1.1}\gain{+4.2} & \textbf{49.5}{\scriptsize$\pm$1.1}\gain{+15.6} \\
DINO-WM PushT (7)     & 45.3{\scriptsize$\pm$3.0} & 57.1{\scriptsize$\pm$4.5}\gain{+11.8} & \textbf{58.8}{\scriptsize$\pm$3.2}\gain{+13.4} & 34.2{\scriptsize$\pm$2.4} & 42.7{\scriptsize$\pm$4.0}\gain{+8.5} & \textbf{44.8}{\scriptsize$\pm$2.5}\gain{+10.6} \\
\midrule
\multicolumn{7}{l}{\textit{Environment shifts (TS)}} \\
PushT recolors (3)    & 38.1{\scriptsize$\pm$1.7} & 40.8{\scriptsize$\pm$1.6}\gain{+2.7} & \textbf{55.2}{\scriptsize$\pm$0.8}\gain{+17.1} & 27.4{\scriptsize$\pm$1.1} & 26.9{\scriptsize$\pm$0.6}\gain{$-$0.5} & \textbf{36.4}{\scriptsize$\pm$0.8}\gain{+9.0} \\
PushObj shapes (3)    & 34.7{\scriptsize$\pm$1.6} & 55.2{\scriptsize$\pm$2.0}\gain{+20.5} & \textbf{58.3}{\scriptsize$\pm$1.9}\gain{+23.6} & 24.5{\scriptsize$\pm$1.0} & 38.2{\scriptsize$\pm$1.7}\gain{+13.7} & \textbf{43.0}{\scriptsize$\pm$1.3}\gain{+18.5} \\
Maze dynamics (2)     & 85.6{\scriptsize$\pm$1.7} & 87.6{\scriptsize$\pm$2.0}\gain{+2.0} & \textbf{89.2}{\scriptsize$\pm$1.4}\gain{+3.6} & 66.5{\scriptsize$\pm$1.1} & 68.2{\scriptsize$\pm$1.3}\gain{+1.7} & \textbf{68.8}{\scriptsize$\pm$1.2}\gain{+2.3} \\
\bottomrule
\end{tabular}
\end{subtable}
\vspace{0.5em}
\begin{subtable}{\textwidth}
\centering
\caption{Predictor and encoder adapted.}
\label{tab:main_enc}
\begin{tabular}{l ccc @{\hspace{9pt}} ccc}
\toprule
 & \multicolumn{3}{c}{Final} & \multicolumn{3}{c}{Mean over replans} \\
\cmidrule(lr){2-4}\cmidrule(lr){5-7}
Setting (shifts) & Frozen & AdaJEPA & OpTTA & Frozen & AdaJEPA & OpTTA \\
\midrule
\multicolumn{7}{l}{\textit{In distribution}} \\
LeWM PushT            & 80.0{\scriptsize$\pm$2.5} & 86.8{\scriptsize$\pm$1.5}\gain{+6.8} & \textbf{90.4}{\scriptsize$\pm$1.7}\gain{+10.4} & 59.1{\scriptsize$\pm$1.1} & 65.6{\scriptsize$\pm$1.9}\gain{+6.5} & \textbf{70.7}{\scriptsize$\pm$2.0}\gain{+11.6} \\
LeWM Reacher          & 98.4{\scriptsize$\pm$0.7} & 99.2{\scriptsize$\pm$0.5}\gain{+0.8} & \textbf{100.0}{\scriptsize$\pm$0.0}\gain{+1.6} & 75.6{\scriptsize$\pm$0.9} & 74.7{\scriptsize$\pm$1.2}\gain{$-$0.9} & \textbf{77.1}{\scriptsize$\pm$0.6}\gain{+1.5} \\
LeWM TwoRoom          & \textbf{93.6}{\scriptsize$\pm$1.3} & 92.4{\scriptsize$\pm$1.0}\gain{$-$1.2} & 91.6{\scriptsize$\pm$0.7}\gain{$-$2.0} & \textbf{80.7}{\scriptsize$\pm$0.8} & 79.0{\scriptsize$\pm$0.8}\gain{$-$1.7} & 79.7{\scriptsize$\pm$0.8}\gain{$-$1.1} \\
TS PushT              & 68.4{\scriptsize$\pm$2.3} & 83.2{\scriptsize$\pm$2.9}\gain{+14.8} & \textbf{90.8}{\scriptsize$\pm$1.6}\gain{+22.4} & 49.4{\scriptsize$\pm$1.2} & 56.8{\scriptsize$\pm$2.7}\gain{+7.4} & \textbf{67.2}{\scriptsize$\pm$0.8}\gain{+17.8} \\
\midrule
\multicolumn{7}{l}{\textit{Visual corruptions}} \\
LeWM PushT (7)        & 58.5{\scriptsize$\pm$0.8} & 63.5{\scriptsize$\pm$0.8}\gain{+5.0} & \textbf{68.4}{\scriptsize$\pm$1.0}\gain{+9.9} & 43.9{\scriptsize$\pm$0.7} & 47.9{\scriptsize$\pm$0.7}\gain{+4.1} & \textbf{53.0}{\scriptsize$\pm$0.8}\gain{+9.2} \\
LeWM Reacher (7)      & 52.8{\scriptsize$\pm$0.5} & 62.3{\scriptsize$\pm$1.1}\gain{+9.5} & \textbf{70.5}{\scriptsize$\pm$1.0}\gain{+17.7} & 38.4{\scriptsize$\pm$0.6} & 41.7{\scriptsize$\pm$0.7}\gain{+3.2} & \textbf{47.1}{\scriptsize$\pm$0.7}\gain{+8.7} \\
LeWM TwoRoom (7)      & \textbf{61.0}{\scriptsize$\pm$1.7} & 57.1{\scriptsize$\pm$1.2}\gain{$-$3.8} & 58.4{\scriptsize$\pm$1.4}\gain{$-$2.6} & \textbf{51.9}{\scriptsize$\pm$1.8} & 49.0{\scriptsize$\pm$1.6}\gain{$-$2.9} & 50.1{\scriptsize$\pm$1.5}\gain{$-$1.8} \\
TS PushT (7)          & 47.8{\scriptsize$\pm$1.5} & 60.7{\scriptsize$\pm$1.9}\gain{+12.9} & \textbf{74.7}{\scriptsize$\pm$1.5}\gain{+26.9} & 34.0{\scriptsize$\pm$1.3} & 39.6{\scriptsize$\pm$1.4}\gain{+5.6} & \textbf{50.8}{\scriptsize$\pm$1.2}\gain{+16.8} \\
\midrule
\multicolumn{7}{l}{\textit{Environment shifts (TS)}} \\
PushT recolors (3)    & 38.1{\scriptsize$\pm$1.7} & 44.5{\scriptsize$\pm$1.3}\gain{+6.4} & \textbf{54.9}{\scriptsize$\pm$1.7}\gain{+16.8} & 27.4{\scriptsize$\pm$1.1} & 29.7{\scriptsize$\pm$1.0}\gain{+2.3} & \textbf{36.9}{\scriptsize$\pm$1.5}\gain{+9.5} \\
PushObj shapes (3)    & 34.7{\scriptsize$\pm$1.6} & 55.5{\scriptsize$\pm$1.1}\gain{+20.8} & \textbf{57.9}{\scriptsize$\pm$2.0}\gain{+23.2} & 24.5{\scriptsize$\pm$1.0} & 38.1{\scriptsize$\pm$1.3}\gain{+13.6} & \textbf{42.5}{\scriptsize$\pm$1.7}\gain{+18.0} \\
Maze dynamics (2)     & 85.6{\scriptsize$\pm$1.7} & 87.6{\scriptsize$\pm$1.7}\gain{+2.0} & \textbf{88.2}{\scriptsize$\pm$1.7}\gain{+2.6} & 66.5{\scriptsize$\pm$1.1} & \textbf{68.0}{\scriptsize$\pm$1.2}\gain{+1.5} & 67.7{\scriptsize$\pm$1.1}\gain{+1.2} \\
\bottomrule
\end{tabular}
\end{subtable}
\end{table}

\paragraph{Across tasks and world models.}
On PushT visual corruptions, the optimistic objective yields about $2.5\times$ the prediction-only adaptation gain in final success for both LeWM and temporal straightening with predictor-only adaptation, and about $2\times$ the gain when the encoder is also adapted.
With DINO-WM, AdaJEPA and OpTTA yield final-success adaptation gains of $11.8$ and $13.4$ points, respectively, on PushT visual corruptions.
On Reacher visual corruptions with LeWM, the ratio is similar: $2.4\times$ with predictor-only adaptation and $1.9\times$ with joint adaptation.
TwoRoom is a notable exception: under visual corruptions, both TTA objectives reduce final success relative to Frozen under both adaptation scopes, a failure mode discussed below.

\paragraph{Environment shifts.}
With temporal straightening, OpTTA yields positive final-success adaptation gains across PushT recolorings, PushObj shape changes, and maze dynamics under both adaptation scopes.
On PushT recolorings, AdaJEPA yields final-success adaptation gains of $2.7$ and $6.4$ points under predictor-only and joint predictor--encoder adaptation, respectively, whereas OpTTA yields gains of $17.1$ and $16.8$ points.
On PushObj shape changes, both TTA objectives yield large and similar gains; on maze dynamics, Frozen already achieves $85.6\%$ final success, leaving limited room for further improvement.

\paragraph{Online sample efficiency.}
In most settings, OpTTA also accumulates success earlier within the replanning budget: its \emph{Mean over replans} is higher in 13 of the 15 shifted settings (per-setting success curves in Appendix~\ref{app:curves}).

\paragraph{In-distribution performance.}
Even without distribution shift, both TTA methods improve final success on PushT for every world model and adaptation scope, by $0.7$--$14.8$ points for AdaJEPA and $4.7$--$23.2$ points for OpTTA.
Reacher is near ceiling for all methods ($98.4$--$100\%$), and on TwoRoom both TTA methods end slightly below Frozen ($93.6\%$), as under shift.

\paragraph{Gradient-based planning.}
With a gradient-based planner, OpTTA's gains over Frozen on temporal-straightening PushT are similar to those with CEM (Appendix~\ref{app:gd}).

\subsection{Ablations and Sensitivity Analysis}
\label{sec:ablations}
\begin{figure}[t]
\centering
\begin{subfigure}[t]{0.31\textwidth}
\centering
\includegraphics[width=\linewidth]{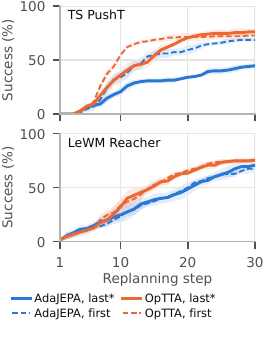}
\caption{Adapted predictor block.}
\label{fig:ablations_params}
\end{subfigure}\hfill
\begin{subfigure}[t]{0.31\textwidth}
\centering
\includegraphics[width=\linewidth]{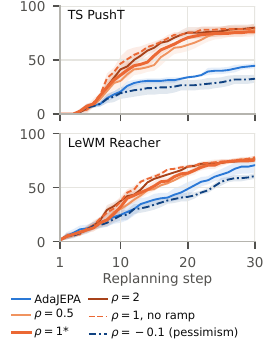}
\caption{Strength and schedule of $\alpha_k$.}
\label{fig:ablations_alpha}
\end{subfigure}\hfill
\begin{subfigure}[t]{0.31\textwidth}
\centering
\includegraphics[width=\linewidth]{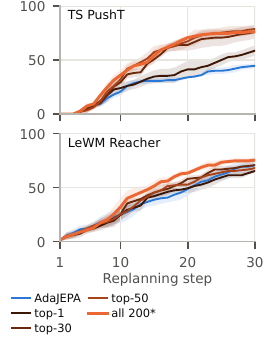}
\caption{Candidate rollouts $\mathcal{P}_\theta$.}
\label{fig:ablations_rollouts}
\end{subfigure}
\caption{Ablations of OpTTA's design choices, varied one at a time (blue: AdaJEPA; orange: OpTTA; $^*$: default). Rows show temporal-straightening PushT under darkening $0.7$ and LeWM Reacher under darkening $0.85$, with the predictor adapted; in (c), top-$n$ keeps the $n$ lowest-cost candidates. Cumulative success is averaged over three seeds of $50$ episodes; shading shows $\pm 1$ s.e.}
\label{fig:ablations}
\end{figure}

We vary OpTTA's three design choices (Section~\ref{sec:method}) one at a time (Figure~\ref{fig:ablations}).

\paragraph{Adapted predictor block.}
Figure~\ref{fig:ablations_params} adapts either the first or the last (default) transformer block of the predictor, with both objectives restricted to that block.
OpTTA's final success changes little with this choice ($76.0\%$ and $72.7\%$ on PushT; $75.3\%$ for both on Reacher) and is higher than AdaJEPA's for both blocks.
AdaJEPA is more sensitive to the block on PushT, where adapting the first block instead of the last raises its final success from $44.7\%$ to $68.7\%$; on Reacher, its two choices reach $70.7\%$ and $67.3\%$.

\paragraph{Strength and schedule of $\alpha_k$.}
Figure~\ref{fig:ablations_alpha} varies $\rho$ over a fourfold range, $\rho\in\{0.5,1,2\}$, and removes the $\sqrt{k/K_{\max}}$ ramp.
OpTTA is fairly robust to $\rho$: final success stays within $76.0$--$80.0\%$ on PushT and $75.3$--$76.7\%$ on Reacher, and mean over replans within $45.2$--$50.9\%$ and $45.7$--$48.9\%$, all well above AdaJEPA ($44.7\%$ and $70.7\%$ final; $26.4\%$ and $38.3\%$ mean over replans).
Reversing the sign of the planning term ($\rho=-0.1$, pessimism) lowers final success below AdaJEPA's on both tasks ($32.7\%$ and $60.7\%$), so the gains come from optimism.
Removing the ramp does not reduce final success ($79.3\%$ and $78.0\%$) and accumulates success earlier (mean over replans of $52.8\%$ and $50.2\%$, versus $47.8\%$ and $46.2\%$ with the ramp).

\paragraph{Candidate rollouts.}
Figure~\ref{fig:ablations_rollouts} restricts $\mathcal{P}_\theta$ to the lowest-cost candidate (top-1) or to the $30$ CEM elites (top-30), instead of all $200$ candidates of CEM's final iteration.
The top-1 candidate gives OpTTA's lowest final success on both tasks ($58.7\%$ on PushT and $65.3\%$ on Reacher), below AdaJEPA's $70.7\%$ on Reacher.
The elites match all candidates in final success on PushT ($76.0\%$ each) but not on Reacher ($70.7\%$ versus $75.3\%$), and all candidates yield the highest mean over replans on both tasks ($47.8\%$ and $46.2\%$, versus $44.4\%$ and $41.6\%$ for the elites).

\paragraph{Computational cost.}
Table~\ref{tab:compute} reports the time of one replanning step, planning plus the adaptation update, on clean PushT.
On LeWM and temporal straightening, OpTTA's update takes about $20$\,ms longer than AdaJEPA's, increasing the time per replan by $5$--$16\%$.
This cost is offset by fewer replans: because OpTTA solves tasks earlier, its compute time per solved task on LeWM and temporal straightening is lower than AdaJEPA's.
On DINO-WM, this update costs nearly as much as planning.
OpTTA evaluates the planning term on the candidate rollouts in batches.
For LeWM and temporal straightening, all $200$ candidates fit in one batch that does not saturate the GPU, so the cost barely depends on $|\mathcal{P}_\theta|$; for DINO-WM, whose larger latent state requires more memory, we process them in four sequential batches of $50$, so restricting $\mathcal{P}_\theta$ to the $30$ CEM elites (a single batch) cuts the cost from $2.3$ to $0.35$\,s per replan.
\providecommand{\gain}[1]{\,{\scriptsize(#1)}}
\begin{table}[t]
\centering
\footnotesize
\setlength{\tabcolsep}{4pt}
\caption{Time per replan (ms) on one RTX PRO 6000 GPU: planning, which does not depend on the method and is averaged over each row, plus the adaptation update (parentheses). OpTTA$_{30}$ restricts $\mathcal{P}_\theta$ to the $30$ CEM elites (Figure~\ref{fig:ablations_rollouts}). LeWM times are wall time and include the environment step; its adaptation time is the difference to Frozen.}
\label{tab:compute}
\begin{tabular}{lcccccc}
\toprule
 & Frozen & AdaJEPA & OpTTA & OpTTA$_{30}$ & AdaJEPA + enc. & OpTTA + enc. \\
\midrule
LeWM PushT    & 118 & 141\gain{+23} & 163\gain{+45} & 165\gain{+47} & 149\gain{+31} & 169\gain{+51} \\
TS PushT      & 379 & 389\gain{+10} & 410\gain{+31} & 410\gain{+31} & 394\gain{+15} & 415\gain{+36} \\
DINO-WM PushT & 2955 & 2971\gain{+16} & 5233\gain{+2278} & 3304\gain{+349} & NA & NA \\
\bottomrule
\end{tabular}
\end{table}

\begin{figure}[!htbp]
\centering
\includegraphics[width=0.94\linewidth]{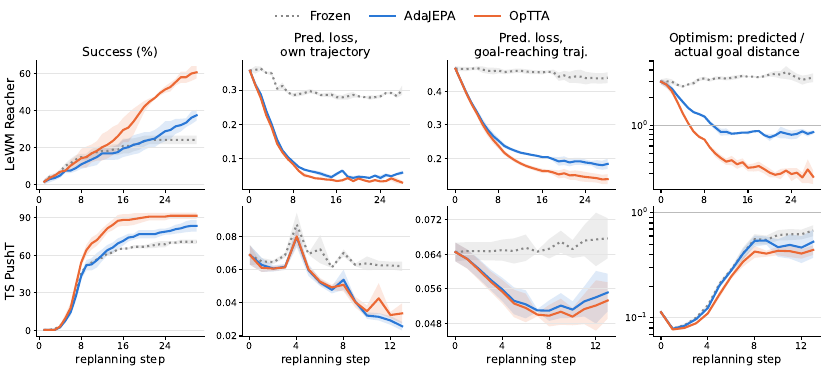}
\caption{Adaptation diagnostics for LeWM Reacher (top, gamma $1.5$) and temporal-straightening PushT (bottom, darkening $0.85$), predictor adapted; mean $\pm$ s.e.\ over three seeds of 50 episodes. Own-trajectory loss is measured on the newest transition before adapting on it; the goal-reaching trajectory is the dataset's start-to-goal trajectory.}
\label{fig:qualitative}
\end{figure}

\paragraph{How optimism changes adaptation.}
Figure~\ref{fig:qualitative} helps explain OpTTA's advantage.
It does not come from fitting the agent's own experience better: OpTTA solves more episodes than AdaJEPA, yet its prediction loss on the transitions the agent executes is not consistently lower.
The difference shows up instead in what the model predicts beyond those transitions.
On Reacher, OpTTA is more accurate than AdaJEPA on a goal-reaching trajectory that the agent never generated; on PushT the two are similar.
On both tasks, and especially on Reacher, OpTTA's model also becomes more optimistic, expecting its chosen plan to close more of the remaining distance to the goal.
This is what optimism is meant to do: planning with an optimistic model either reaches the goal or produces transitions that the prediction loss then uses to correct the model.

\FloatBarrier

\section{Limitations}
On TwoRoom, neither AdaJEPA nor OpTTA improves over the pretrained model, and the best parameters to update vary across tasks (Figure~\ref{fig:ablations_params}).
Understanding this task dependence remains open.
OpTTA also adds computation per replan, although its higher mean success indicates that it often solves tasks earlier.
The performance--cost trade-off warrants further study, and Table~\ref{tab:compute} and Figure~\ref{fig:ablations} show that the default configuration of OpTTA need not be optimal, despite the demonstrated adaptation gain.

\section{Conclusion}
\label{sec:conclusion}
We introduce OpTTA, which augments prediction-based test-time adaptation with an optimistic planning loss. Across three JEPA world models, both adaptation scopes, visual and environment shifts, and CEM and gradient-based planners, OpTTA improves final success over prediction-only adaptation in most settings. Our design of the optimism weight $\alpha_k$ is practical but not necessarily optimal, and its optimal design remains an open question. More broadly, our results suggest that world-model adaptation should account for the downstream planning objective rather than prediction alone, and that optimism offers a simple way to do so.

\clearpage
\bibliography{final}
\bibliographystyle{conference}

\appendix
\addtocontents{toc}{\protect\setcounter{tocdepth}{3}}
\newpage
\setcounter{tocdepth}{3}
\tableofcontents

\makeatletter
\let\app@origfigure\figure
\renewcommand{\figure}[1][]{\app@origfigure[!htbp]}
\let\app@origtable\table
\renewcommand{\table}[1][]{\app@origtable[!htbp]}
\setlength{\@fptop}{0pt}
\setlength{\@fpbot}{0pt plus 1fil}
\makeatother
\renewcommand{\topfraction}{0.95}
\renewcommand{\bottomfraction}{0.9}
\renewcommand{\textfraction}{0.05}
\renewcommand{\floatpagefraction}{0.85}
\setcounter{topnumber}{3}
\setcounter{bottomnumber}{2}
\setcounter{totalnumber}{4}
\setlength{\floatsep}{8pt plus 2pt minus 2pt}
\setlength{\textfloatsep}{10pt plus 2pt minus 3pt}
\setlength{\intextsep}{8pt plus 2pt minus 2pt}

\section{Additional experimental results}
\label{app:experiments}

This appendix provides a detailed account of the experimental setup and results supporting the comparisons in Section~\ref{sec:experiments}.

\subsection{Experimental details}
\label{app:setup}

\paragraph{Evaluation unit.}
For visual shifts, a task--shift combination is one world-model--task pair evaluated under one corruption type after averaging over the severity levels used for that checkpoint. Seven corruptions for each of LeWM PushT, LeWM Reacher, LeWM TwoRoom, temporal-straightening PushT, and DINO-WM PushT give 35 visual combinations. Eight environment-specific combinations additionally evaluate three object recolorings, three held-out object shapes, and two dynamics changes, yielding 43 task--shift combinations in total.

\paragraph{Models, tasks, and shifts.}
We evaluate LeWM, temporal straightening (TS), and DINO-WM on the tasks described in Section~\ref{sec:setup}. Visual shifts comprise blur, salt-and-pepper noise, darkening, gamma, hue, contrast, and image translation. Environment-specific shifts change object color or geometry, mass, or damping. LeWM and TS results average over four severity levels selected separately for each checkpoint using only the frozen model. DINO-WM uses one level per corruption in the aggregate comparison.

\paragraph{Implementation.}
Our OpTTA implementation builds on the public \href{https://github.com/agentic-learning-ai-lab/adajepa}{AdaJEPA codebase}. We retain its test-time adaptation pipeline and add the optimistic planning loss and gradient-balancing rule. The LeWM experiments are implemented in the public \href{https://github.com/lucas-maes/le-wm}{LeWM codebase} using a faithful port of AdaJEPA's adaptation procedure. All pretrained checkpoints used in our experiments are also publicly available.

\paragraph{Controlled comparison.}
Every comparison uses the same pretrained checkpoint, planner, start--goal pairs, transition buffer, adapted parameter set, and update schedule. The frozen model receives no parameter updates. AdaJEPA estimates the model using $\mathcal{L}_{\rm pred}$, whereas OpTTA uses the optimistic, task-aware model-estimation objective in \eqref{eq:optta_objective}. Their comparison therefore isolates the adaptation objective. All primary comparisons use $\rho=1$ in \eqref{eq:alpha}, without tuning it for individual models, planners, tasks, shifts, or severity levels.

\paragraph{Planning and adaptation protocol.}
CEM plans five actions, executes the first, and replans for at most 30 steps. Adaptation is task-local: each task begins from the pretrained weights and an empty transition buffer. LeWM and TS conditions use five seeds with 50 paired start--goal episodes per seed. DINO-WM uses three seeds with 50 episodes per seed. DINO-WM's encoder remains frozen; for LeWM and TS, we report both predictor-only and joint predictor--encoder adaptation.

\paragraph{Hyperparameters.}
Table~\ref{app:tab:hparams} lists the planning and adaptation hyperparameters of every run. Adaptation follows AdaJEPA \citep{wang2026adajepa}: at every replan, one Adam step on the last predictor block, using a buffer of the five most recent transitions (learning rates per model in the table). AdaJEPA and OpTTA share all of these settings. OpTTA adds the base optimism $\rho$, the ramp $\sqrt{k/K_{\max}}$, the stabilizer $\varepsilon$, and the choice of $\mathcal{P}_\theta$ and $\Omega_f$, and these are the same across world models, tasks, shifts, and severity levels.
\begin{table}[h]
\centering
\small
\setlength{\tabcolsep}{4pt}
\renewcommand{\arraystretch}{1.05}
\caption{Hyperparameters of the planning and adaptation protocol. Adaptation follows AdaJEPA \citep{wang2026adajepa} and is identical for AdaJEPA and OpTTA; OpTTA adds only the optimism hyperparameters, which are the same for every world model, task, and shift. A latent step covers five environment steps (frameskip).}
\label{app:tab:hparams}
\begin{tabular}{l >{\centering\arraybackslash}p{3.2cm} >{\centering\arraybackslash}p{2.7cm} >{\centering\arraybackslash}p{2.7cm}}
\toprule
Hyperparameter & LeWM & TS & DINO-WM \\
\midrule
\multicolumn{4}{l}{\textit{Planning (CEM; all methods)}} \\
Candidates per iteration $M$ & $200$ & $200$ & $200$ \\
Elites & $30$ & $30$ & $30$ \\
CEM iterations & $10$ & $10$ & $10$ \\
Planning horizon & \multicolumn{3}{c}{$5$ latent steps ($25$ environment steps)} \\
Executed per replan & \multicolumn{3}{c}{first latent step ($5$ environment steps)} \\
Replanning budget $K_{\max}$ & $30$ & $30$ & $30$ \\
Warm start & \multicolumn{3}{c}{unexecuted part of the previous plan} \\
Numerical precision & float32 (TF32 off) & float32 & bfloat16\textsuperscript{b} \\
\midrule
\multicolumn{4}{l}{\textit{Adaptation (AdaJEPA and OpTTA)}} \\
Optimizer & \multicolumn{3}{c}{Adam, re-initialized at every update} \\
Gradient steps per replan $U$ & $1$ & $1$ & $1$ \\
Transition buffer $N$ & \multicolumn{3}{c}{$5$ most recent transitions} \\
Adapted predictor parameters & last block, final norm, output projection & last block, final norm & last block, final norm \\
Predictor learning rate $\eta$ & $5\times10^{-5}$ & $5\times10^{-4}$ & $5\times10^{-4}$ \\
Adapted encoder parameters\textsuperscript{a} & projector (BatchNorm in eval mode) & projection head & --- (frozen) \\
Encoder learning rate & $5\times10^{-5}$ & $10^{-5}$ & --- \\
\midrule
\multicolumn{4}{l}{\textit{Optimism (OpTTA only)}} \\
Base optimism $\rho$ & $1$ & $1$ & $1$ \\
Ramp in \eqref{eq:alpha} & \multicolumn{3}{c}{$\sqrt{k/K_{\max}}$ with $K_{\max}=30$} \\
Stabilizer $\varepsilon$ & $10^{-12}$ & $10^{-12}$ & $10^{-12}$ \\
Candidate rollouts $\mathcal{P}_\theta$ & \multicolumn{3}{c}{all $M=200$ candidates of CEM's final iteration} \\
Planning-loss parameters $\Omega_f$ & \multicolumn{3}{c}{adapted predictor parameters (never the encoder)} \\
\midrule
\multicolumn{4}{l}{\textit{Gradient-descent planner (TS only; Appendix~\ref{app:gd})}} \\
Action optimizer & \multicolumn{3}{c}{Adam, learning rate $0.1$, $100$ steps, zero initialization} \\
Horizon / executed per replan & \multicolumn{3}{c}{$5$ / $1$ latent steps, as for CEM} \\
Replanning budget $K_{\max}$ & \multicolumn{3}{c}{$20$} \\
Optimism & \multicolumn{3}{c}{$\rho=1$ without the ramp; $\mathcal{P}_\theta$ is the optimized plan} \\
\bottomrule
\end{tabular}

\vspace{2pt}
{\footnotesize \textsuperscript{a}Joint predictor--encoder scope only; the predictor-only scope adapts no encoder parameters.\\
\textsuperscript{b}Autocast for CEM's cost evaluation (all methods) and the forward pass of $\mathcal{L}_{\rm plan}$ (OpTTA); the plan and the $\mathcal{L}_{\rm pred}$ update stay float32.}
\end{table}

\paragraph{Metrics and aggregation.}
\emph{Final} is the percentage of episodes solved by replan 30, and \emph{Mean over replans} is the average cumulative success rate across replans 1 through 30. Within each seed, we first average episodes at each severity and then average severity levels within a corruption; reported means and standard errors are computed across seeds. The visual-corruption rows in Table~\ref{tab:main} additionally average the seven corruptions within each world-model--task setting, giving each corruption equal weight. The tables also report each adaptation method's gain over Frozen, paired by seed within the same adaptation scope.

\subsection{Per-shift results}
\label{app:pershift}
Tables~\ref{app:tab:pershift_pred} and~\ref{app:tab:pershift_enc} disaggregate the shifted results in Table~\ref{tab:main} by corruption type or environment change. The predictor-only table contains all 43 task--shift combinations; the joint predictor--encoder table contains the 36 combinations evaluated with LeWM and temporal straightening, since the DINO-WM encoder remains frozen.
\providecommand{\gain}[1]{\,{\scriptsize(#1)}}
\begin{table}[p]
\centering
\scriptsize
\setlength{\tabcolsep}{1.6pt}
\renewcommand{\arraystretch}{0.94}
\caption{Predictor adapted. Planning success (\%, mean $\pm$ s.e.\ over seeds, five seeds unless noted) per shift, in the layout of Table~\ref{tab:main}. Each visual corruption averages the severity levels listed in its label. Parentheses give the gain over Frozen, paired by seed; bold marks the best method at the printed precision.}
\label{app:tab:pershift_pred}
\begin{tabular}{l ccc @{\hspace{5pt}} ccc}
\toprule
 & \multicolumn{3}{c}{Final} & \multicolumn{3}{c}{Mean over replans} \\
\cmidrule(lr){2-4}\cmidrule(lr){5-7}
Setting & Frozen & AdaJEPA & OpTTA & Frozen & AdaJEPA & OpTTA \\
\midrule
\multicolumn{7}{l}{\textit{LeWM PushT}} \\
\quad Blur (2, 4, 6, 8) & 77.6{\scriptsize$\pm$1.3} & 78.7{\scriptsize$\pm$1.8}\gain{+1.1} & \textbf{85.5}{\scriptsize$\pm$1.1}\gain{+7.9} & 58.7{\scriptsize$\pm$0.7} & 60.3{\scriptsize$\pm$1.1}\gain{+1.6} & \textbf{67.4}{\scriptsize$\pm$0.7}\gain{+8.7} \\
\quad Salt \& pepper (0.025, 0.05, 0.075, 0.1) & 53.5{\scriptsize$\pm$1.4} & 53.9{\scriptsize$\pm$0.9}\gain{+0.4} & \textbf{60.5}{\scriptsize$\pm$1.3}\gain{+7.0} & 40.5{\scriptsize$\pm$1.0} & 41.5{\scriptsize$\pm$0.7}\gain{+1.1} & \textbf{46.5}{\scriptsize$\pm$0.8}\gain{+6.0} \\
\quad Darkening (0.95, 0.9, 0.85, 0.8) & 54.8{\scriptsize$\pm$1.2} & 58.2{\scriptsize$\pm$1.7}\gain{+3.4} & \textbf{61.8}{\scriptsize$\pm$0.9}\gain{+7.0} & 41.8{\scriptsize$\pm$0.8} & 44.8{\scriptsize$\pm$1.2}\gain{+2.9} & \textbf{48.2}{\scriptsize$\pm$0.7}\gain{+6.4} \\
\quad Gamma (0.5, 0.75, 1.25, 1.5) & 76.7{\scriptsize$\pm$1.7} & 82.6{\scriptsize$\pm$0.5}\gain{+5.9} & \textbf{87.8}{\scriptsize$\pm$1.0}\gain{+11.1} & 57.5{\scriptsize$\pm$1.2} & 62.4{\scriptsize$\pm$0.7}\gain{+4.8} & \textbf{68.4}{\scriptsize$\pm$0.8}\gain{+10.9} \\
\quad Hue (10, 20, 30, 40) & 44.1{\scriptsize$\pm$2.6} & 46.6{\scriptsize$\pm$1.7}\gain{+2.5} & \textbf{52.4}{\scriptsize$\pm$1.9}\gain{+8.3} & 31.6{\scriptsize$\pm$2.1} & 34.1{\scriptsize$\pm$1.4}\gain{+2.5} & \textbf{39.1}{\scriptsize$\pm$1.3}\gain{+7.5} \\
\quad Contrast (0.5, 0.75, 1.25, 1.5) & 43.5{\scriptsize$\pm$1.4} & 51.4{\scriptsize$\pm$2.2}\gain{+7.9} & \textbf{58.3}{\scriptsize$\pm$0.9}\gain{+14.8} & 32.2{\scriptsize$\pm$1.1} & 38.5{\scriptsize$\pm$1.9}\gain{+6.4} & \textbf{43.4}{\scriptsize$\pm$1.0}\gain{+11.2} \\
\quad Translation (10, 20, 30, 40) & 59.5{\scriptsize$\pm$1.1} & 65.6{\scriptsize$\pm$0.7}\gain{+6.1} & \textbf{73.4}{\scriptsize$\pm$0.4}\gain{+13.9} & 44.8{\scriptsize$\pm$1.0} & 49.7{\scriptsize$\pm$0.5}\gain{+4.9} & \textbf{57.0}{\scriptsize$\pm$0.3}\gain{+12.3} \\
\midrule
\multicolumn{7}{l}{\textit{LeWM Reacher}} \\
\quad Blur (2, 4, 6, 8) & 52.5{\scriptsize$\pm$1.6} & 58.0{\scriptsize$\pm$1.6}\gain{+5.5} & \textbf{69.2}{\scriptsize$\pm$0.7}\gain{+16.7} & 38.2{\scriptsize$\pm$0.8} & 40.1{\scriptsize$\pm$0.6}\gain{+1.9} & \textbf{46.5}{\scriptsize$\pm$0.8}\gain{+8.3} \\
\quad Salt \& pepper (0.025, 0.05, 0.075, 0.1) & 84.6{\scriptsize$\pm$0.8} & 84.3{\scriptsize$\pm$0.6}\gain{$-$0.3} & \textbf{88.9}{\scriptsize$\pm$0.6}\gain{+4.3} & 61.5{\scriptsize$\pm$0.9} & 60.1{\scriptsize$\pm$0.6}\gain{$-$1.5} & \textbf{64.9}{\scriptsize$\pm$0.7}\gain{+3.3} \\
\quad Darkening (0.95, 0.9, 0.85, 0.8) & 61.2{\scriptsize$\pm$0.9} & 73.8{\scriptsize$\pm$0.5}\gain{+12.6} & \textbf{82.5}{\scriptsize$\pm$0.5}\gain{+21.3} & 45.4{\scriptsize$\pm$0.6} & 48.8{\scriptsize$\pm$0.9}\gain{+3.4} & \textbf{55.4}{\scriptsize$\pm$0.8}\gain{+10.0} \\
\quad Gamma (0.5, 0.75, 1.25, 1.5) & 22.5{\scriptsize$\pm$1.6} & 36.8{\scriptsize$\pm$1.2}\gain{+14.3} & \textbf{54.5}{\scriptsize$\pm$1.1}\gain{+32.0} & 16.5{\scriptsize$\pm$1.3} & 21.6{\scriptsize$\pm$0.9}\gain{+5.2} & \textbf{30.9}{\scriptsize$\pm$1.3}\gain{+14.4} \\
\quad Hue (10, 20, 30, 40) & 56.5{\scriptsize$\pm$1.5} & 63.5{\scriptsize$\pm$2.2}\gain{+7.0} & \textbf{72.8}{\scriptsize$\pm$2.4}\gain{+16.3} & 39.6{\scriptsize$\pm$1.0} & 42.1{\scriptsize$\pm$1.2}\gain{+2.5} & \textbf{48.7}{\scriptsize$\pm$1.4}\gain{+9.1} \\
\quad Contrast (0.5, 0.75, 1.25, 1.5) & 69.0{\scriptsize$\pm$1.1} & 75.3{\scriptsize$\pm$1.0}\gain{+6.3} & \textbf{85.0}{\scriptsize$\pm$0.9}\gain{+16.0} & 51.3{\scriptsize$\pm$1.1} & 52.8{\scriptsize$\pm$1.4}\gain{+1.5} & \textbf{59.7}{\scriptsize$\pm$1.2}\gain{+8.4} \\
\quad Translation (10, 20, 30, 40) & 23.2{\scriptsize$\pm$1.9} & 30.5{\scriptsize$\pm$1.4}\gain{+7.3} & \textbf{42.7}{\scriptsize$\pm$2.9}\gain{+19.5} & 16.6{\scriptsize$\pm$1.3} & 18.9{\scriptsize$\pm$0.5}\gain{+2.3} & \textbf{24.3}{\scriptsize$\pm$1.5}\gain{+7.7} \\
\midrule
\multicolumn{7}{l}{\textit{LeWM TwoRoom}} \\
\quad Blur (0.5, 0.75, 1, 2) & \textbf{68.7}{\scriptsize$\pm$2.4} & 63.1{\scriptsize$\pm$2.3}\gain{$-$5.6} & 62.5{\scriptsize$\pm$2.9}\gain{$-$6.2} & \textbf{58.5}{\scriptsize$\pm$2.5} & 54.5{\scriptsize$\pm$2.3}\gain{$-$4.0} & 54.3{\scriptsize$\pm$2.8}\gain{$-$4.2} \\
\quad Salt \& pepper (0.005, 0.01, 0.015, 0.025) & \textbf{61.2}{\scriptsize$\pm$1.7} & 56.6{\scriptsize$\pm$2.2}\gain{$-$4.6} & 56.7{\scriptsize$\pm$1.8}\gain{$-$4.5} & \textbf{50.9}{\scriptsize$\pm$2.2} & 48.1{\scriptsize$\pm$1.9}\gain{$-$2.7} & 47.7{\scriptsize$\pm$1.8}\gain{$-$3.1} \\
\quad Darkening (0.2, 0.5, 0.7, 0.8) & \textbf{55.2}{\scriptsize$\pm$3.2} & 50.0{\scriptsize$\pm$2.1}\gain{$-$5.2} & 49.9{\scriptsize$\pm$2.1}\gain{$-$5.3} & \textbf{47.3}{\scriptsize$\pm$2.8} & 43.8{\scriptsize$\pm$2.1}\gain{$-$3.5} & 43.8{\scriptsize$\pm$1.9}\gain{$-$3.5} \\
\quad Gamma (0.1, 0.15, 0.2, 0.3) & \textbf{79.2}{\scriptsize$\pm$1.2} & 77.1{\scriptsize$\pm$1.4}\gain{$-$2.1} & 77.7{\scriptsize$\pm$0.7}\gain{$-$1.5} & \textbf{66.4}{\scriptsize$\pm$0.8} & 65.4{\scriptsize$\pm$1.1}\gain{$-$1.1} & 65.4{\scriptsize$\pm$1.1}\gain{$-$1.0} \\
\quad Hue (30, 40, 50, 60) & 64.6{\scriptsize$\pm$2.3} & \textbf{65.3}{\scriptsize$\pm$2.8}\gain{+0.7} & 62.3{\scriptsize$\pm$1.9}\gain{$-$2.3} & \textbf{55.8}{\scriptsize$\pm$2.3} & 53.5{\scriptsize$\pm$2.3}\gain{$-$2.4} & 53.7{\scriptsize$\pm$1.7}\gain{$-$2.1} \\
\quad Contrast (0.2, 0.4, 0.5, 0.6) & \textbf{57.4}{\scriptsize$\pm$3.7} & 52.1{\scriptsize$\pm$2.4}\gain{$-$5.3} & 53.2{\scriptsize$\pm$1.9}\gain{$-$4.2} & \textbf{49.4}{\scriptsize$\pm$3.4} & 46.2{\scriptsize$\pm$2.3}\gain{$-$3.2} & 46.5{\scriptsize$\pm$2.1}\gain{$-$2.8} \\
\quad Translation (10, 20, 30, 40) & \textbf{40.6}{\scriptsize$\pm$1.0} & 39.1{\scriptsize$\pm$0.9}\gain{$-$1.5} & 38.3{\scriptsize$\pm$1.3}\gain{$-$2.3} & \textbf{35.1}{\scriptsize$\pm$0.8} & 34.0{\scriptsize$\pm$1.2}\gain{$-$1.2} & 33.4{\scriptsize$\pm$1.4}\gain{$-$1.7} \\
\midrule
\multicolumn{7}{l}{\textit{TS PushT}} \\
\quad Blur (1, 2, 4, 6) & 49.0{\scriptsize$\pm$1.7} & 67.3{\scriptsize$\pm$1.1}\gain{+18.3} & \textbf{81.6}{\scriptsize$\pm$0.8}\gain{+32.6} & 34.6{\scriptsize$\pm$1.5} & 43.7{\scriptsize$\pm$1.3}\gain{+9.1} & \textbf{55.0}{\scriptsize$\pm$0.7}\gain{+20.4} \\
\quad Salt \& pepper (0.025, 0.05, 0.075, 0.1) & 47.4{\scriptsize$\pm$2.2} & 65.2{\scriptsize$\pm$1.8}\gain{+17.8} & \textbf{86.7}{\scriptsize$\pm$1.8}\gain{+39.3} & 32.8{\scriptsize$\pm$1.7} & 41.1{\scriptsize$\pm$1.4}\gain{+8.4} & \textbf{56.8}{\scriptsize$\pm$1.4}\gain{+24.1} \\
\quad Darkening (0.65, 0.7, 0.75, 0.8) & 50.5{\scriptsize$\pm$2.2} & 50.6{\scriptsize$\pm$1.7}\gain{+0.1} & \textbf{68.5}{\scriptsize$\pm$1.1}\gain{+18.0} & 35.1{\scriptsize$\pm$1.6} & 33.2{\scriptsize$\pm$1.1}\gain{$-$1.9} & \textbf{45.9}{\scriptsize$\pm$1.2}\gain{+10.8} \\
\quad Gamma (0.3, 0.7, 2, 4) & 44.0{\scriptsize$\pm$2.0} & 56.0{\scriptsize$\pm$0.8}\gain{+12.0} & \textbf{75.9}{\scriptsize$\pm$2.5}\gain{+31.9} & 31.9{\scriptsize$\pm$1.5} & 35.5{\scriptsize$\pm$0.5}\gain{+3.6} & \textbf{48.6}{\scriptsize$\pm$1.7}\gain{+16.7} \\
\quad Hue (10, 20, 30, 40) & 43.4{\scriptsize$\pm$2.5} & 52.3{\scriptsize$\pm$2.4}\gain{+8.9} & \textbf{59.5}{\scriptsize$\pm$2.8}\gain{+16.1} & 32.3{\scriptsize$\pm$2.3} & 35.5{\scriptsize$\pm$1.9}\gain{+3.2} & \textbf{41.3}{\scriptsize$\pm$2.1}\gain{+9.0} \\
\quad Contrast (0.4, 0.8, 2, 4) & 54.2{\scriptsize$\pm$1.7} & 62.6{\scriptsize$\pm$1.2}\gain{+8.4} & \textbf{80.7}{\scriptsize$\pm$1.5}\gain{+26.5} & 37.7{\scriptsize$\pm$1.3} & 41.6{\scriptsize$\pm$1.1}\gain{+3.9} & \textbf{55.1}{\scriptsize$\pm$1.5}\gain{+17.4} \\
\quad Translation (10, 20, 30, 40) & 46.4{\scriptsize$\pm$1.9} & 53.8{\scriptsize$\pm$3.3}\gain{+7.4} & \textbf{65.0}{\scriptsize$\pm$2.9}\gain{+18.6} & 33.4{\scriptsize$\pm$1.6} & 36.5{\scriptsize$\pm$2.1}\gain{+3.1} & \textbf{44.0}{\scriptsize$\pm$2.2}\gain{+10.6} \\
\midrule
\multicolumn{7}{l}{\textit{DINO-WM PushT (one level, three seeds)}} \\
\quad Blur (2) & 66.0{\scriptsize$\pm$6.1} & 78.0{\scriptsize$\pm$5.0}\gain{+12.0} & \textbf{78.7}{\scriptsize$\pm$5.3}\gain{+12.7} & 47.9{\scriptsize$\pm$4.1} & 58.8{\scriptsize$\pm$6.2}\gain{+10.8} & \textbf{60.4}{\scriptsize$\pm$4.5}\gain{+12.5} \\
\quad Salt \& pepper (0.001) & 44.0{\scriptsize$\pm$2.3} & \textbf{69.3}{\scriptsize$\pm$2.9}\gain{+25.3} & 65.3{\scriptsize$\pm$3.3}\gain{+21.3} & 33.9{\scriptsize$\pm$2.6} & 47.5{\scriptsize$\pm$2.6}\gain{+13.6} & \textbf{49.8}{\scriptsize$\pm$3.1}\gain{+15.9} \\
\quad Darkening (0.1) & 20.7{\scriptsize$\pm$4.7} & 27.3{\scriptsize$\pm$2.9}\gain{+6.7} & \textbf{31.3}{\scriptsize$\pm$1.8}\gain{+10.7} & 15.8{\scriptsize$\pm$3.7} & 19.0{\scriptsize$\pm$1.6}\gain{+3.2} & \textbf{23.4}{\scriptsize$\pm$1.3}\gain{+7.6} \\
\quad Gamma (0.2) & 52.7{\scriptsize$\pm$2.9} & 68.0{\scriptsize$\pm$7.0}\gain{+15.3} & \textbf{72.7}{\scriptsize$\pm$5.7}\gain{+20.0} & 38.6{\scriptsize$\pm$2.5} & 52.4{\scriptsize$\pm$5.4}\gain{+13.8} & \textbf{55.0}{\scriptsize$\pm$3.8}\gain{+16.4} \\
\quad Hue (180) & 76.7{\scriptsize$\pm$5.8} & \textbf{77.3}{\scriptsize$\pm$6.8}\gain{+0.7} & 76.0{\scriptsize$\pm$5.3}\gain{$-$0.7} & 58.4{\scriptsize$\pm$5.3} & \textbf{60.2}{\scriptsize$\pm$5.6}\gain{+1.8} & 59.4{\scriptsize$\pm$4.0}\gain{+1.0} \\
\quad Contrast (2) & 17.3{\scriptsize$\pm$2.4} & 27.3{\scriptsize$\pm$4.8}\gain{+10.0} & \textbf{30.7}{\scriptsize$\pm$2.4}\gain{+13.3} & 13.8{\scriptsize$\pm$2.9} & 19.8{\scriptsize$\pm$3.2}\gain{+6.0} & \textbf{22.6}{\scriptsize$\pm$2.1}\gain{+8.9} \\
\quad Translation (5) & 40.0{\scriptsize$\pm$4.6} & 52.7{\scriptsize$\pm$8.1}\gain{+12.7} & \textbf{56.7}{\scriptsize$\pm$2.4}\gain{+16.7} & 30.9{\scriptsize$\pm$3.2} & 41.0{\scriptsize$\pm$7.0}\gain{+10.1} & \textbf{42.6}{\scriptsize$\pm$2.6}\gain{+11.7} \\
\midrule
\multicolumn{7}{l}{\textit{TS environment shifts}} \\
\quad PushT, red agent & 53.6{\scriptsize$\pm$2.5} & 68.8{\scriptsize$\pm$3.7}\gain{+15.2} & \textbf{85.6}{\scriptsize$\pm$2.5}\gain{+32.0} & 40.9{\scriptsize$\pm$2.1} & 46.7{\scriptsize$\pm$2.7}\gain{+5.9} & \textbf{60.6}{\scriptsize$\pm$1.8}\gain{+19.7} \\
\quad PushT, red anchor & 44.8{\scriptsize$\pm$2.8} & 42.8{\scriptsize$\pm$3.6}\gain{$-$2.0} & \textbf{62.0}{\scriptsize$\pm$2.1}\gain{+17.2} & 30.1{\scriptsize$\pm$2.1} & 26.0{\scriptsize$\pm$2.5}\gain{$-$4.1} & \textbf{37.1}{\scriptsize$\pm$1.8}\gain{+7.0} \\
\quad PushT, red block & 16.0{\scriptsize$\pm$2.2} & 10.8{\scriptsize$\pm$2.6}\gain{$-$5.2} & \textbf{18.0}{\scriptsize$\pm$3.0}\gain{+2.0} & 11.2{\scriptsize$\pm$1.1} & 8.0{\scriptsize$\pm$1.9}\gain{$-$3.2} & \textbf{11.4}{\scriptsize$\pm$1.5}\gain{+0.3} \\
\quad PushObj, shape i & 27.6{\scriptsize$\pm$2.5} & 50.4{\scriptsize$\pm$3.4}\gain{+22.8} & \textbf{54.0}{\scriptsize$\pm$3.8}\gain{+26.4} & 19.5{\scriptsize$\pm$2.0} & 34.8{\scriptsize$\pm$2.5}\gain{+15.3} & \textbf{38.8}{\scriptsize$\pm$2.7}\gain{+19.3} \\
\quad PushObj, shape small t & 53.2{\scriptsize$\pm$3.9} & 75.2{\scriptsize$\pm$2.8}\gain{+22.0} & \textbf{78.0}{\scriptsize$\pm$3.0}\gain{+24.8} & 37.1{\scriptsize$\pm$2.4} & 51.9{\scriptsize$\pm$2.3}\gain{+14.7} & \textbf{58.2}{\scriptsize$\pm$2.0}\gain{+21.1} \\
\quad PushObj, shape square & 23.2{\scriptsize$\pm$1.7} & 40.0{\scriptsize$\pm$2.6}\gain{+16.8} & \textbf{42.8}{\scriptsize$\pm$3.4}\gain{+19.6} & 16.9{\scriptsize$\pm$1.1} & 27.9{\scriptsize$\pm$2.4}\gain{+11.0} & \textbf{31.9}{\scriptsize$\pm$2.5}\gain{+15.1} \\
\quad Medium maze, low mass & 90.8{\scriptsize$\pm$1.9} & 92.0{\scriptsize$\pm$1.8}\gain{+1.2} & \textbf{93.6}{\scriptsize$\pm$1.8}\gain{+2.8} & 74.5{\scriptsize$\pm$1.7} & 74.7{\scriptsize$\pm$1.6}\gain{+0.3} & \textbf{75.3}{\scriptsize$\pm$2.1}\gain{+0.9} \\
\quad Medium maze, high damping & 80.4{\scriptsize$\pm$2.4} & 83.2{\scriptsize$\pm$2.6}\gain{+2.8} & \textbf{84.8}{\scriptsize$\pm$2.0}\gain{+4.4} & 58.6{\scriptsize$\pm$1.4} & 61.7{\scriptsize$\pm$1.5}\gain{+3.1} & \textbf{62.3}{\scriptsize$\pm$0.8}\gain{+3.7} \\
\bottomrule
\end{tabular}
\end{table}

\begin{table}[p]
\centering
\scriptsize
\setlength{\tabcolsep}{1.6pt}
\renewcommand{\arraystretch}{0.94}
\caption{Predictor and encoder adapted (DINO-WM has no encoder variant). Planning success (\%, mean $\pm$ s.e.\ over seeds, five seeds unless noted) per shift, in the layout of Table~\ref{tab:main}. Each visual corruption averages the severity levels listed in its label. Parentheses give the gain over Frozen, paired by seed; bold marks the best method at the printed precision.}
\label{app:tab:pershift_enc}
\begin{tabular}{l ccc @{\hspace{5pt}} ccc}
\toprule
 & \multicolumn{3}{c}{Final} & \multicolumn{3}{c}{Mean over replans} \\
\cmidrule(lr){2-4}\cmidrule(lr){5-7}
Setting & Frozen & AdaJEPA & OpTTA & Frozen & AdaJEPA & OpTTA \\
\midrule
\multicolumn{7}{l}{\textit{LeWM PushT}} \\
\quad Blur (2, 4, 6, 8) & 77.6{\scriptsize$\pm$1.3} & 80.0{\scriptsize$\pm$1.6}\gain{+2.4} & \textbf{86.5}{\scriptsize$\pm$1.0}\gain{+8.9} & 58.7{\scriptsize$\pm$0.7} & 60.4{\scriptsize$\pm$1.3}\gain{+1.7} & \textbf{67.6}{\scriptsize$\pm$0.7}\gain{+8.9} \\
\quad Salt \& pepper (0.025, 0.05, 0.075, 0.1) & 53.5{\scriptsize$\pm$1.4} & 53.5{\scriptsize$\pm$1.0}\gain{+0.0} & \textbf{58.3}{\scriptsize$\pm$2.7}\gain{+4.8} & 40.5{\scriptsize$\pm$1.0} & 41.3{\scriptsize$\pm$1.1}\gain{+0.8} & \textbf{45.2}{\scriptsize$\pm$1.9}\gain{+4.7} \\
\quad Darkening (0.95, 0.9, 0.85, 0.8) & 54.8{\scriptsize$\pm$1.2} & 59.5{\scriptsize$\pm$2.4}\gain{+4.7} & \textbf{63.9}{\scriptsize$\pm$1.8}\gain{+9.1} & 41.8{\scriptsize$\pm$0.8} & 45.4{\scriptsize$\pm$1.8}\gain{+3.6} & \textbf{50.1}{\scriptsize$\pm$1.6}\gain{+8.3} \\
\quad Gamma (0.5, 0.75, 1.25, 1.5) & 76.7{\scriptsize$\pm$1.7} & 81.6{\scriptsize$\pm$1.4}\gain{+4.9} & \textbf{87.7}{\scriptsize$\pm$0.9}\gain{+11.0} & 57.5{\scriptsize$\pm$1.2} & 61.7{\scriptsize$\pm$1.3}\gain{+4.1} & \textbf{68.5}{\scriptsize$\pm$0.9}\gain{+10.9} \\
\quad Hue (10, 20, 30, 40) & 44.1{\scriptsize$\pm$2.6} & 49.1{\scriptsize$\pm$1.7}\gain{+5.0} & \textbf{52.3}{\scriptsize$\pm$2.1}\gain{+8.2} & 31.6{\scriptsize$\pm$2.1} & 36.1{\scriptsize$\pm$0.9}\gain{+4.5} & \textbf{39.9}{\scriptsize$\pm$1.4}\gain{+8.4} \\
\quad Contrast (0.5, 0.75, 1.25, 1.5) & 43.5{\scriptsize$\pm$1.4} & 53.4{\scriptsize$\pm$1.8}\gain{+9.9} & \textbf{57.7}{\scriptsize$\pm$1.1}\gain{+14.2} & 32.2{\scriptsize$\pm$1.1} & 39.8{\scriptsize$\pm$1.9}\gain{+7.6} & \textbf{43.8}{\scriptsize$\pm$1.0}\gain{+11.6} \\
\quad Translation (10, 20, 30, 40) & 59.5{\scriptsize$\pm$1.1} & 67.5{\scriptsize$\pm$0.8}\gain{+8.0} & \textbf{72.4}{\scriptsize$\pm$1.3}\gain{+12.9} & 44.8{\scriptsize$\pm$1.0} & 51.0{\scriptsize$\pm$0.7}\gain{+6.2} & \textbf{56.3}{\scriptsize$\pm$1.0}\gain{+11.5} \\
\midrule
\multicolumn{7}{l}{\textit{LeWM Reacher}} \\
\quad Blur (2, 4, 6, 8) & 52.5{\scriptsize$\pm$1.6} & 58.2{\scriptsize$\pm$0.8}\gain{+5.7} & \textbf{71.2}{\scriptsize$\pm$1.8}\gain{+18.7} & 38.2{\scriptsize$\pm$0.8} & 40.2{\scriptsize$\pm$0.7}\gain{+2.1} & \textbf{47.8}{\scriptsize$\pm$0.9}\gain{+9.6} \\
\quad Salt \& pepper (0.025, 0.05, 0.075, 0.1) & 84.6{\scriptsize$\pm$0.8} & 86.9{\scriptsize$\pm$0.9}\gain{+2.3} & \textbf{89.8}{\scriptsize$\pm$1.5}\gain{+5.2} & 61.5{\scriptsize$\pm$0.9} & 62.1{\scriptsize$\pm$1.2}\gain{+0.5} & \textbf{64.9}{\scriptsize$\pm$1.1}\gain{+3.4} \\
\quad Darkening (0.95, 0.9, 0.85, 0.8) & 61.2{\scriptsize$\pm$0.9} & 76.3{\scriptsize$\pm$1.2}\gain{+15.1} & \textbf{80.6}{\scriptsize$\pm$0.3}\gain{+19.4} & 45.4{\scriptsize$\pm$0.6} & 50.1{\scriptsize$\pm$1.0}\gain{+4.6} & \textbf{55.0}{\scriptsize$\pm$0.8}\gain{+9.5} \\
\quad Gamma (0.5, 0.75, 1.25, 1.5) & 22.5{\scriptsize$\pm$1.6} & 40.8{\scriptsize$\pm$2.1}\gain{+18.3} & \textbf{47.6}{\scriptsize$\pm$1.3}\gain{+25.1} & 16.5{\scriptsize$\pm$1.3} & 23.7{\scriptsize$\pm$1.6}\gain{+7.2} & \textbf{27.9}{\scriptsize$\pm$0.6}\gain{+11.5} \\
\quad Hue (10, 20, 30, 40) & 56.5{\scriptsize$\pm$1.5} & 63.1{\scriptsize$\pm$1.6}\gain{+6.6} & \textbf{73.7}{\scriptsize$\pm$2.9}\gain{+17.2} & 39.6{\scriptsize$\pm$1.0} & 42.5{\scriptsize$\pm$0.6}\gain{+2.9} & \textbf{48.7}{\scriptsize$\pm$1.5}\gain{+9.1} \\
\quad Contrast (0.5, 0.75, 1.25, 1.5) & 69.0{\scriptsize$\pm$1.1} & 79.3{\scriptsize$\pm$1.7}\gain{+10.3} & \textbf{88.0}{\scriptsize$\pm$1.2}\gain{+19.0} & 51.3{\scriptsize$\pm$1.1} & 54.2{\scriptsize$\pm$0.5}\gain{+2.9} & \textbf{61.7}{\scriptsize$\pm$0.7}\gain{+10.4} \\
\quad Translation (10, 20, 30, 40) & 23.2{\scriptsize$\pm$1.9} & 31.5{\scriptsize$\pm$2.1}\gain{+8.3} & \textbf{42.3}{\scriptsize$\pm$3.6}\gain{+19.1} & 16.6{\scriptsize$\pm$1.3} & 19.0{\scriptsize$\pm$1.4}\gain{+2.4} & \textbf{24.0}{\scriptsize$\pm$2.2}\gain{+7.4} \\
\midrule
\multicolumn{7}{l}{\textit{LeWM TwoRoom}} \\
\quad Blur (0.5, 0.75, 1, 2) & \textbf{68.7}{\scriptsize$\pm$2.4} & 59.4{\scriptsize$\pm$2.6}\gain{$-$9.3} & 60.2{\scriptsize$\pm$2.3}\gain{$-$8.5} & \textbf{58.5}{\scriptsize$\pm$2.5} & 51.4{\scriptsize$\pm$2.5}\gain{$-$7.1} & 52.2{\scriptsize$\pm$2.3}\gain{$-$6.3} \\
\quad Salt \& pepper (0.005, 0.01, 0.015, 0.025) & \textbf{61.2}{\scriptsize$\pm$1.7} & 56.8{\scriptsize$\pm$1.4}\gain{$-$4.4} & 59.7{\scriptsize$\pm$2.7}\gain{$-$1.5} & \textbf{50.9}{\scriptsize$\pm$2.2} & 48.1{\scriptsize$\pm$2.1}\gain{$-$2.8} & 49.7{\scriptsize$\pm$2.3}\gain{$-$1.1} \\
\quad Darkening (0.2, 0.5, 0.7, 0.8) & \textbf{55.2}{\scriptsize$\pm$3.2} & 51.1{\scriptsize$\pm$1.2}\gain{$-$4.1} & 51.2{\scriptsize$\pm$1.7}\gain{$-$4.0} & \textbf{47.3}{\scriptsize$\pm$2.8} & 44.0{\scriptsize$\pm$1.4}\gain{$-$3.3} & 44.9{\scriptsize$\pm$1.8}\gain{$-$2.4} \\
\quad Gamma (0.1, 0.15, 0.2, 0.3) & \textbf{79.2}{\scriptsize$\pm$1.2} & 78.0{\scriptsize$\pm$1.4}\gain{$-$1.2} & \textbf{79.2}{\scriptsize$\pm$1.5}\gain{+0.0} & 66.4{\scriptsize$\pm$0.8} & 65.5{\scriptsize$\pm$1.3}\gain{$-$0.9} & \textbf{66.9}{\scriptsize$\pm$1.0}\gain{+0.5} \\
\quad Hue (30, 40, 50, 60) & \textbf{64.6}{\scriptsize$\pm$2.3} & 61.9{\scriptsize$\pm$1.8}\gain{$-$2.7} & 62.0{\scriptsize$\pm$1.7}\gain{$-$2.6} & \textbf{55.8}{\scriptsize$\pm$2.3} & 53.3{\scriptsize$\pm$2.0}\gain{$-$2.5} & 53.5{\scriptsize$\pm$1.7}\gain{$-$2.3} \\
\quad Contrast (0.2, 0.4, 0.5, 0.6) & \textbf{57.4}{\scriptsize$\pm$3.7} & 50.2{\scriptsize$\pm$1.7}\gain{$-$7.2} & 52.0{\scriptsize$\pm$1.4}\gain{$-$5.4} & \textbf{49.4}{\scriptsize$\pm$3.4} & 43.8{\scriptsize$\pm$1.7}\gain{$-$5.5} & 45.5{\scriptsize$\pm$1.5}\gain{$-$3.9} \\
\quad Translation (10, 20, 30, 40) & 40.6{\scriptsize$\pm$1.0} & 42.6{\scriptsize$\pm$1.6}\gain{+2.0} & \textbf{44.4}{\scriptsize$\pm$1.9}\gain{+3.8} & 35.1{\scriptsize$\pm$0.8} & 36.7{\scriptsize$\pm$1.6}\gain{+1.6} & \textbf{37.8}{\scriptsize$\pm$1.9}\gain{+2.7} \\
\midrule
\multicolumn{7}{l}{\textit{TS PushT}} \\
\quad Blur (1, 2, 4, 6) & 49.0{\scriptsize$\pm$1.7} & 69.8{\scriptsize$\pm$1.5}\gain{+20.8} & \textbf{81.3}{\scriptsize$\pm$1.4}\gain{+32.3} & 34.6{\scriptsize$\pm$1.5} & 46.0{\scriptsize$\pm$1.2}\gain{+11.5} & \textbf{56.8}{\scriptsize$\pm$0.9}\gain{+22.2} \\
\quad Salt \& pepper (0.025, 0.05, 0.075, 0.1) & 47.4{\scriptsize$\pm$2.2} & 69.8{\scriptsize$\pm$2.3}\gain{+22.4} & \textbf{86.5}{\scriptsize$\pm$2.0}\gain{+39.1} & 32.8{\scriptsize$\pm$1.7} & 44.1{\scriptsize$\pm$2.0}\gain{+11.4} & \textbf{58.7}{\scriptsize$\pm$1.6}\gain{+26.0} \\
\quad Darkening (0.65, 0.7, 0.75, 0.8) & 50.5{\scriptsize$\pm$2.2} & 51.5{\scriptsize$\pm$2.9}\gain{+1.0} & \textbf{68.7}{\scriptsize$\pm$2.7}\gain{+18.2} & 35.1{\scriptsize$\pm$1.6} & 32.6{\scriptsize$\pm$1.8}\gain{$-$2.5} & \textbf{45.9}{\scriptsize$\pm$2.1}\gain{+10.8} \\
\quad Gamma (0.3, 0.7, 2, 4) & 44.0{\scriptsize$\pm$2.0} & 60.7{\scriptsize$\pm$2.2}\gain{+16.7} & \textbf{76.6}{\scriptsize$\pm$1.9}\gain{+32.6} & 31.9{\scriptsize$\pm$1.5} & 38.9{\scriptsize$\pm$1.0}\gain{+7.0} & \textbf{50.0}{\scriptsize$\pm$1.3}\gain{+18.0} \\
\quad Hue (10, 20, 30, 40) & 43.4{\scriptsize$\pm$2.5} & 52.7{\scriptsize$\pm$2.4}\gain{+9.3} & \textbf{60.8}{\scriptsize$\pm$3.4}\gain{+17.4} & 32.3{\scriptsize$\pm$2.3} & 36.0{\scriptsize$\pm$2.1}\gain{+3.8} & \textbf{42.4}{\scriptsize$\pm$2.5}\gain{+10.1} \\
\quad Contrast (0.4, 0.8, 2, 4) & 54.2{\scriptsize$\pm$1.7} & 63.2{\scriptsize$\pm$1.8}\gain{+9.0} & \textbf{81.7}{\scriptsize$\pm$1.3}\gain{+27.5} & 37.7{\scriptsize$\pm$1.3} & 40.8{\scriptsize$\pm$1.5}\gain{+3.1} & \textbf{55.2}{\scriptsize$\pm$1.2}\gain{+17.6} \\
\quad Translation (10, 20, 30, 40) & 46.4{\scriptsize$\pm$1.9} & 57.3{\scriptsize$\pm$2.7}\gain{+10.9} & \textbf{67.3}{\scriptsize$\pm$2.9}\gain{+20.9} & 33.4{\scriptsize$\pm$1.6} & 38.6{\scriptsize$\pm$2.1}\gain{+5.2} & \textbf{46.6}{\scriptsize$\pm$2.4}\gain{+13.2} \\
\midrule
\multicolumn{7}{l}{\textit{TS environment shifts}} \\
\quad PushT, red agent & 53.6{\scriptsize$\pm$2.5} & 77.6{\scriptsize$\pm$3.1}\gain{+24.0} & \textbf{87.6}{\scriptsize$\pm$1.9}\gain{+34.0} & 40.9{\scriptsize$\pm$2.1} & 52.9{\scriptsize$\pm$2.5}\gain{+12.1} & \textbf{64.4}{\scriptsize$\pm$2.0}\gain{+23.5} \\
\quad PushT, red anchor & 44.8{\scriptsize$\pm$2.8} & 46.0{\scriptsize$\pm$3.3}\gain{+1.2} & \textbf{61.6}{\scriptsize$\pm$2.1}\gain{+16.8} & 30.1{\scriptsize$\pm$2.1} & 28.9{\scriptsize$\pm$2.0}\gain{$-$1.2} & \textbf{36.3}{\scriptsize$\pm$1.5}\gain{+6.1} \\
\quad PushT, red block & \textbf{16.0}{\scriptsize$\pm$2.2} & 10.0{\scriptsize$\pm$1.1}\gain{$-$6.0} & 15.6{\scriptsize$\pm$2.4}\gain{$-$0.4} & \textbf{11.2}{\scriptsize$\pm$1.1} & 7.2{\scriptsize$\pm$1.0}\gain{$-$3.9} & 10.1{\scriptsize$\pm$2.0}\gain{$-$1.1} \\
\quad PushObj, shape i & 27.6{\scriptsize$\pm$2.5} & \textbf{53.6}{\scriptsize$\pm$3.5}\gain{+26.0} & \textbf{53.6}{\scriptsize$\pm$3.2}\gain{+26.0} & 19.5{\scriptsize$\pm$2.0} & 37.1{\scriptsize$\pm$2.1}\gain{+17.6} & \textbf{38.9}{\scriptsize$\pm$2.7}\gain{+19.3} \\
\quad PushObj, shape small t & 53.2{\scriptsize$\pm$3.9} & 72.8{\scriptsize$\pm$3.4}\gain{+19.6} & \textbf{77.6}{\scriptsize$\pm$2.9}\gain{+24.4} & 37.1{\scriptsize$\pm$2.4} & 49.5{\scriptsize$\pm$2.6}\gain{+12.4} & \textbf{57.4}{\scriptsize$\pm$2.6}\gain{+20.3} \\
\quad PushObj, shape square & 23.2{\scriptsize$\pm$1.7} & 40.0{\scriptsize$\pm$3.8}\gain{+16.8} & \textbf{42.4}{\scriptsize$\pm$3.3}\gain{+19.2} & 16.9{\scriptsize$\pm$1.1} & 27.6{\scriptsize$\pm$1.8}\gain{+10.8} & \textbf{31.3}{\scriptsize$\pm$2.3}\gain{+14.4} \\
\quad Medium maze, low mass & 90.8{\scriptsize$\pm$1.9} & 91.2{\scriptsize$\pm$1.5}\gain{+0.4} & \textbf{92.4}{\scriptsize$\pm$1.8}\gain{+1.6} & \textbf{74.5}{\scriptsize$\pm$1.7} & 74.3{\scriptsize$\pm$2.3}\gain{$-$0.2} & \textbf{74.5}{\scriptsize$\pm$1.8}\gain{$-$0.0} \\
\quad Medium maze, high damping & 80.4{\scriptsize$\pm$2.4} & \textbf{84.0}{\scriptsize$\pm$2.4}\gain{+3.6} & \textbf{84.0}{\scriptsize$\pm$1.9}\gain{+3.6} & 58.6{\scriptsize$\pm$1.4} & \textbf{61.7}{\scriptsize$\pm$1.1}\gain{+3.1} & 60.9{\scriptsize$\pm$1.0}\gain{+2.3} \\
\bottomrule
\end{tabular}
\end{table}

\FloatBarrier

\subsection{Success across replanning steps}
\label{app:curves}
\begin{figure}[p]
\centering
\includegraphics[width=\textwidth]{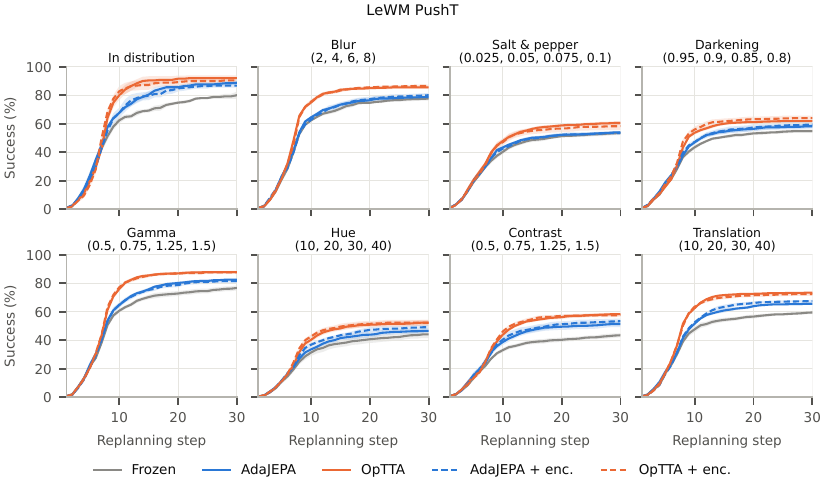}
\caption{LeWM PushT: cumulative success versus replanning step, in distribution and under each visual corruption, averaged over the severity levels in the panel title. Curves are means over seeds with $\pm1$ s.e.\ bands; solid lines adapt the predictor, dashed lines the predictor and encoder. The mean of each curve is the table's mean over replans, and its last point the table's Final.}
\label{app:fig:lewm_pusht}
\end{figure}

\begin{figure}[p]
\centering
\includegraphics[width=\textwidth]{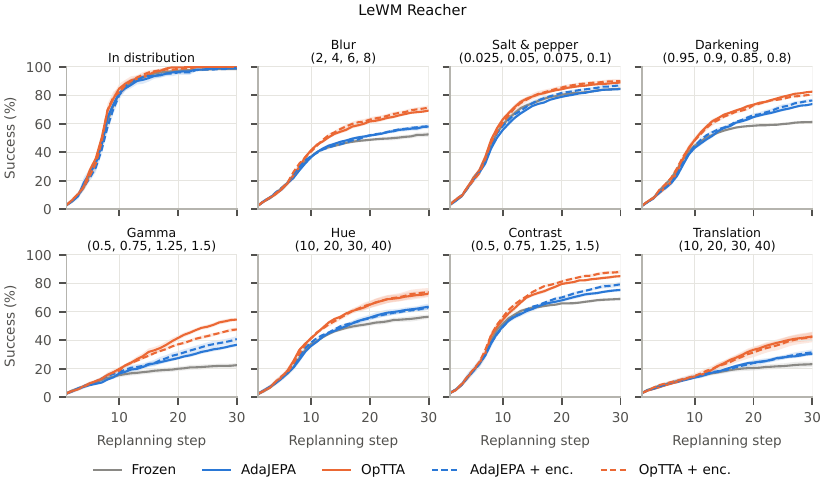}
\caption{LeWM Reacher: cumulative success versus replanning step, in distribution and under each visual corruption, averaged over the severity levels in the panel title. Curves are means over seeds with $\pm1$ s.e.\ bands; solid lines adapt the predictor, dashed lines the predictor and encoder. The mean of each curve is the table's mean over replans, and its last point the table's Final.}
\label{app:fig:lewm_reacher}
\end{figure}

\begin{figure}[p]
\centering
\includegraphics[width=\textwidth]{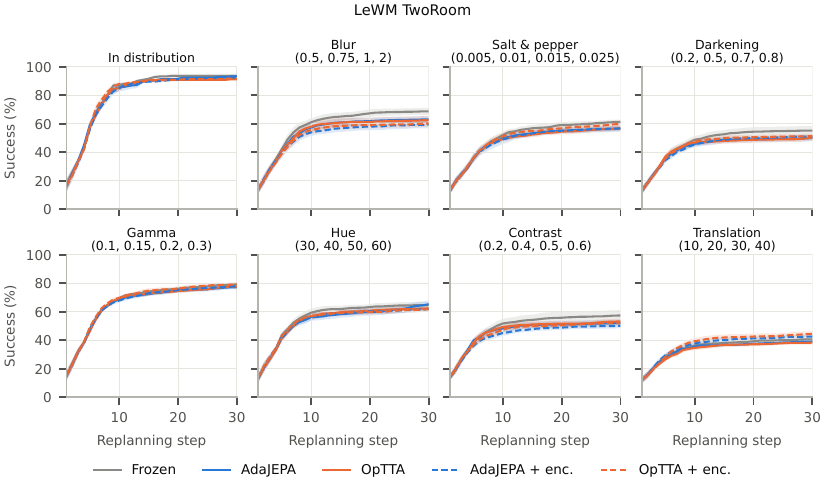}
\caption{LeWM TwoRoom: cumulative success versus replanning step, in distribution and under each visual corruption, averaged over the severity levels in the panel title. Curves are means over seeds with $\pm1$ s.e.\ bands; solid lines adapt the predictor, dashed lines the predictor and encoder. The mean of each curve is the table's mean over replans, and its last point the table's Final.}
\label{app:fig:lewm_tworoom}
\end{figure}

\begin{figure}[p]
\centering
\includegraphics[width=\textwidth]{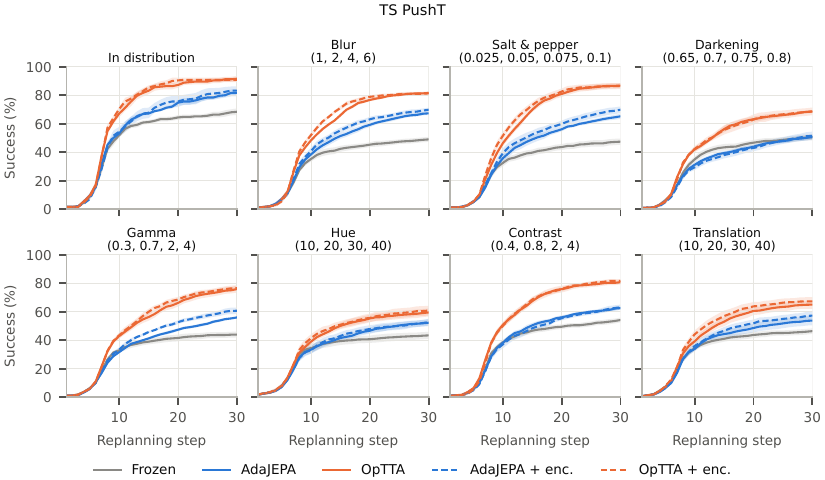}
\caption{TS PushT: cumulative success versus replanning step, in distribution and under each visual corruption, averaged over the severity levels in the panel title. Curves are means over seeds with $\pm1$ s.e.\ bands; solid lines adapt the predictor, dashed lines the predictor and encoder. The mean of each curve is the table's mean over replans, and its last point the table's Final.}
\label{app:fig:ts_pusht}
\end{figure}

\begin{figure}[p]
\centering
\includegraphics[width=\textwidth]{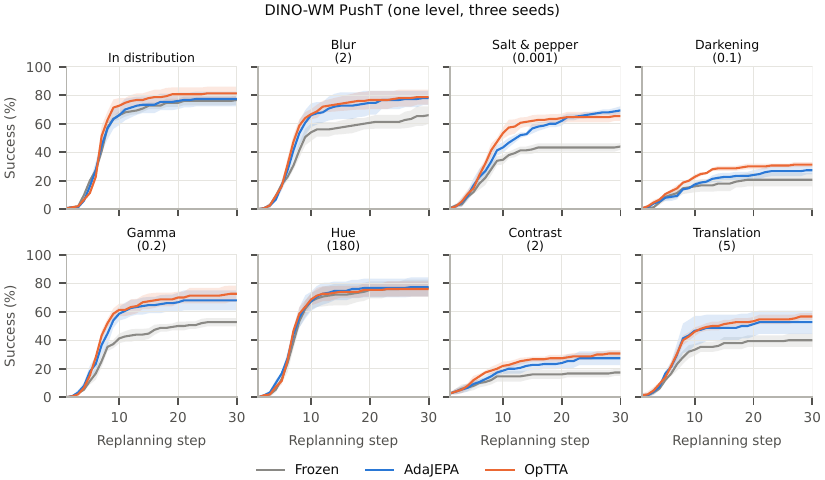}
\caption{DINO-WM PushT (one level, three seeds): cumulative success versus replanning step, in distribution and under each visual corruption, averaged over the severity levels in the panel title. Curves are means over seeds with $\pm1$ s.e.\ bands; all lines adapt only the predictor. The mean of each curve is the table's mean over replans, and its last point the table's Final.}
\label{app:fig:dino_pusht}
\end{figure}

\begin{figure}[p]
\centering
\includegraphics[width=\textwidth]{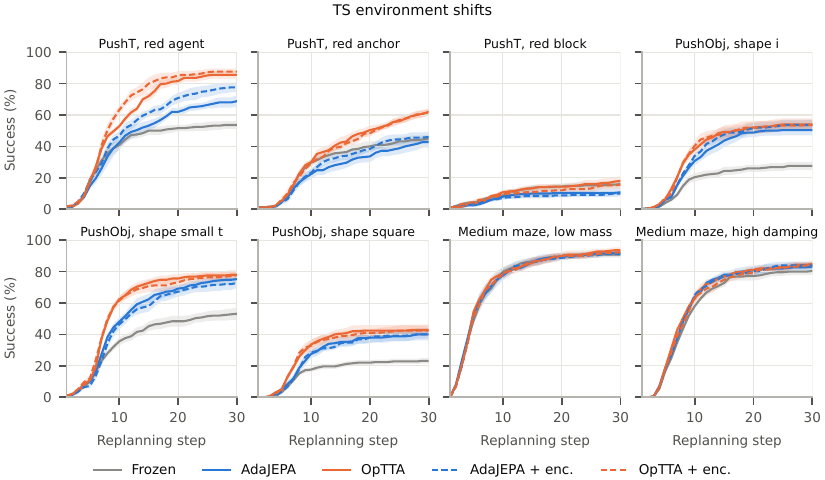}
\caption{TS environment shifts: cumulative success versus replanning step, each environment shift. Curves are means over seeds with $\pm1$ s.e.\ bands; solid lines adapt the predictor, dashed lines the predictor and encoder. The mean of each curve is the table's mean over replans, and its last point the table's Final.}
\label{app:fig:ts_env}
\end{figure}

Figures~\ref{app:fig:lewm_pusht}--\ref{app:fig:ts_env} report cumulative success at every replanning step for each setting. For visual corruptions, curves average the severity levels used in the corresponding table row; environment shifts are shown individually. The final point of each curve is \emph{Final}, and the mean of its 30 points is \emph{Mean over replans}.

\FloatBarrier

\subsection{Success across severity levels}
\label{app:severity}
\begin{figure}[p]
\centering
\includegraphics[width=\textwidth]{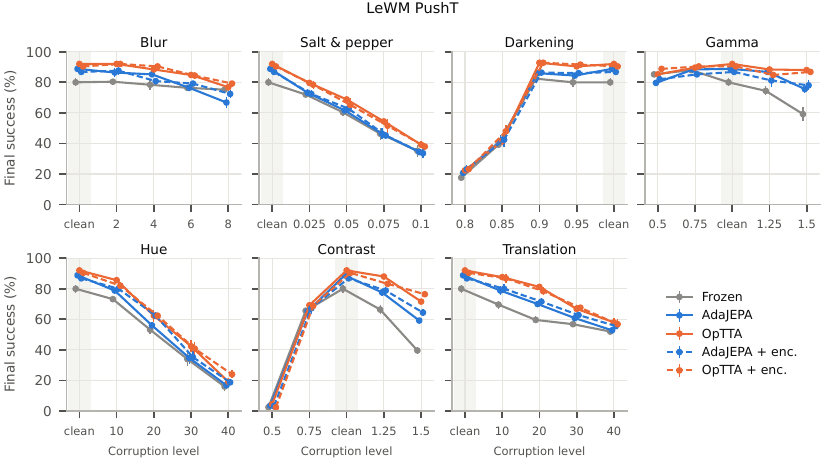}
\caption{LeWM PushT: final success at each level of each visual corruption, with the uncorrupted condition (clean, shaded) placed at the identity value; levels are ordered by value. Points are means over five seeds with $\pm1$ s.e.\ bars; solid lines adapt the predictor, dashed lines the predictor and encoder. The mean over a panel's four corrupted levels is the corresponding entry of Tables~\ref{app:tab:pershift_pred} and~\ref{app:tab:pershift_enc}.}
\label{app:fig:severity_lewm_pusht}
\end{figure}

\begin{figure}[p]
\centering
\includegraphics[width=\textwidth]{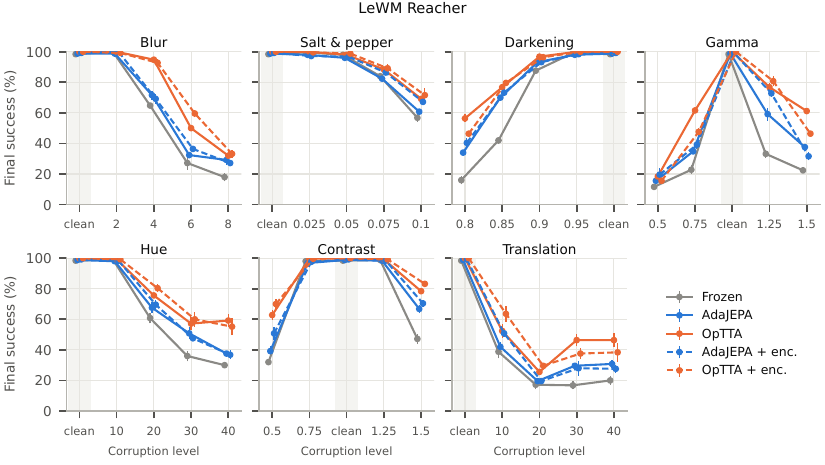}
\caption{LeWM Reacher: final success at each level of each visual corruption, with the uncorrupted condition (clean, shaded) placed at the identity value; levels are ordered by value. Points are means over five seeds with $\pm1$ s.e.\ bars; solid lines adapt the predictor, dashed lines the predictor and encoder. The mean over a panel's four corrupted levels is the corresponding entry of Tables~\ref{app:tab:pershift_pred} and~\ref{app:tab:pershift_enc}.}
\label{app:fig:severity_lewm_reacher}
\end{figure}

\begin{figure}[p]
\centering
\includegraphics[width=\textwidth]{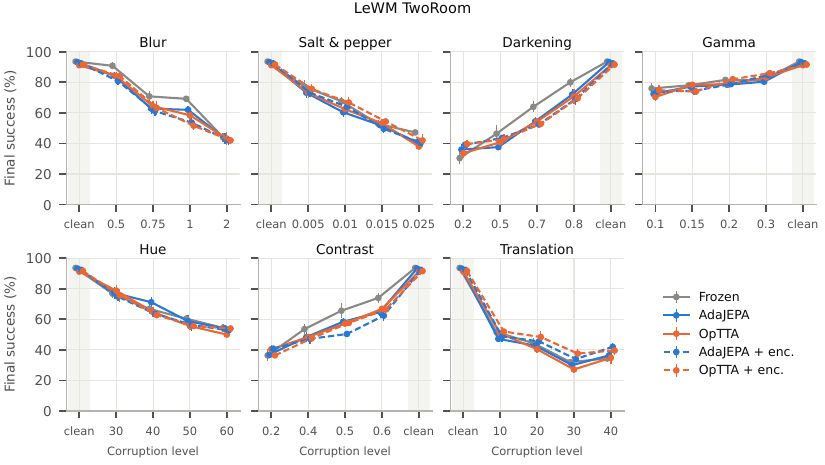}
\caption{LeWM TwoRoom: final success at each level of each visual corruption, with the uncorrupted condition (clean, shaded) placed at the identity value; levels are ordered by value. Points are means over five seeds with $\pm1$ s.e.\ bars; solid lines adapt the predictor, dashed lines the predictor and encoder. The mean over a panel's four corrupted levels is the corresponding entry of Tables~\ref{app:tab:pershift_pred} and~\ref{app:tab:pershift_enc}.}
\label{app:fig:severity_lewm_tworoom}
\end{figure}

\begin{figure}[p]
\centering
\includegraphics[width=\textwidth]{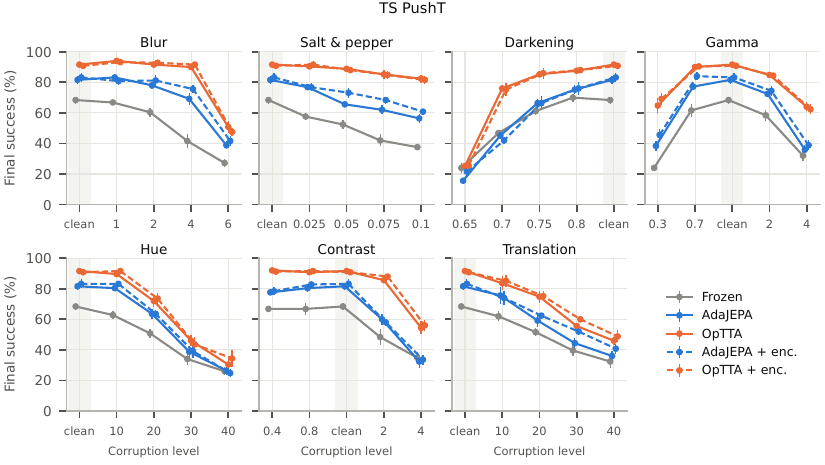}
\caption{TS PushT: final success at each level of each visual corruption, with the uncorrupted condition (clean, shaded) placed at the identity value; levels are ordered by value. Points are means over five seeds with $\pm1$ s.e.\ bars; solid lines adapt the predictor, dashed lines the predictor and encoder. The mean over a panel's four corrupted levels is the corresponding entry of Tables~\ref{app:tab:pershift_pred} and~\ref{app:tab:pershift_enc}.}
\label{app:fig:severity_ts_pusht}
\end{figure}

Figures~\ref{app:fig:severity_lewm_pusht}--\ref{app:fig:severity_ts_pusht} report final success separately at each visual-corruption severity for LeWM and temporal straightening. The clean condition is shown only as a reference and is not included in the corrupted-level averages. DINO-WM is omitted because its aggregate evaluation uses one severity level per corruption.

\FloatBarrier

\subsection{Gradient-based planner}
\label{app:gd}
\begin{figure}[p]
\centering
\includegraphics[width=\textwidth]{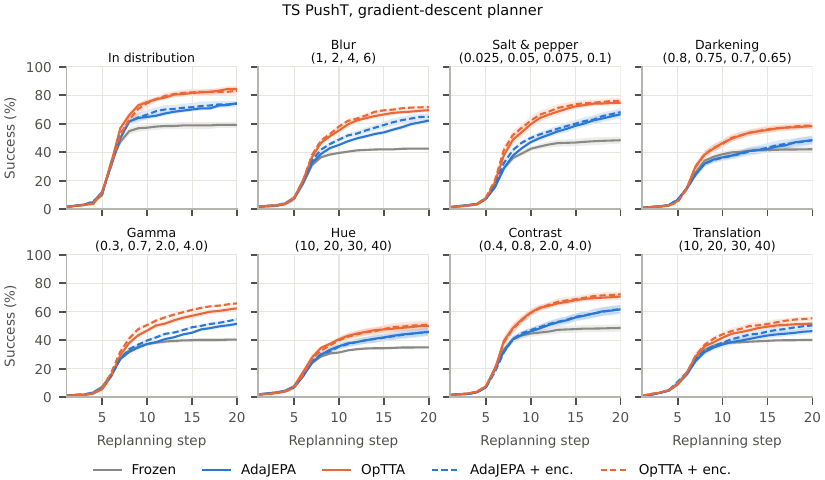}
\caption{Temporal straightening on PushT with a gradient-descent planner: cumulative success versus replanning step over the $20$-replan budget, one panel per row of Table~\ref{app:tab:gd}. Visual corruptions average the severity levels in the panel title. Curves are means over seeds with $\pm1$ s.e.\ bands; solid lines adapt the predictor, dashed lines the predictor and encoder. The mean of each curve is the table's mean over replans, and its last point the table's Final.}
\label{app:fig:gd}
\end{figure}

We evaluate temporal-straightening PushT under a separate gradient-based planning protocol. The planner optimizes the action sequence directly, and $\mathcal{L}_{\rm plan}$ is evaluated on the optimized plan. These runs use 20 replans and constant $\rho=1$ without the optimism ramp. Within this protocol, Frozen, AdaJEPA, and OpTTA use the same start--goal pairs and applicable adaptation settings. Thus, the comparison tests the adaptation objectives under gradient-based planning, but does not isolate planner choice relative to CEM. Table~\ref{app:tab:gd} reports Final at replan 20 and Mean over replans across steps 1--20, and Figure~\ref{app:fig:gd} shows the corresponding success curves.
\providecommand{\gain}[1]{\,{\scriptsize(#1)}}
\begin{table}[h]
\centering
\small
\setlength{\tabcolsep}{2.8pt}
\renewcommand{\arraystretch}{0.96}
\caption{Temporal straightening on PushT with a gradient-descent planner. Planning success (\%, mean $\pm$ s.e.\ over five seeds of $50$ episodes). The gradient-descent runs use a budget of $20$ replans, so Final is success by replanning step $20$ and Mean over replans is the average cumulative success rate across replanning steps $1$--$20$; values are not directly comparable to the $30$-replan Table~\ref{tab:main}. Parentheses report the adaptation gain over Frozen, paired by seed and computed before rounding. Each visual corruption averages its four severity levels.}
\label{app:tab:gd}
\begin{subtable}{\textwidth}
\centering
\caption{Predictor adapted.}
\begin{tabular}{l ccc @{\hspace{9pt}} ccc}
\toprule
 & \multicolumn{3}{c}{Final} & \multicolumn{3}{c}{Mean over replans} \\
\cmidrule(lr){2-4}\cmidrule(lr){5-7}
Setting & Frozen & AdaJEPA & OpTTA & Frozen & AdaJEPA & OpTTA \\
\midrule
In distribution           & 59.2{\scriptsize$\pm$2.1} & 74.4{\scriptsize$\pm$1.2}\gain{+15.2} & \textbf{84.4}{\scriptsize$\pm$1.9}\gain{+25.2} & 42.9{\scriptsize$\pm$1.5} & 49.9{\scriptsize$\pm$0.6}\gain{+6.9} & \textbf{57.0}{\scriptsize$\pm$1.0}\gain{+14.1} \\
\midrule
\multicolumn{7}{l}{\textit{Visual corruptions}} \\
\quad Blur                & 42.5{\scriptsize$\pm$0.8} & 62.2{\scriptsize$\pm$1.3}\gain{+19.7} & \textbf{69.5}{\scriptsize$\pm$1.2}\gain{+27.0} & 30.0{\scriptsize$\pm$0.8} & 37.4{\scriptsize$\pm$1.1}\gain{+7.4} & \textbf{44.4}{\scriptsize$\pm$1.1}\gain{+14.4} \\
\quad Salt \& pepper      & 48.4{\scriptsize$\pm$2.4} & 66.6{\scriptsize$\pm$2.7}\gain{+18.2} & \textbf{74.7}{\scriptsize$\pm$1.8}\gain{+26.3} & 32.4{\scriptsize$\pm$1.4} & 38.9{\scriptsize$\pm$1.3}\gain{+6.5} & \textbf{47.4}{\scriptsize$\pm$1.5}\gain{+15.0} \\
\quad Darkening           & 42.1{\scriptsize$\pm$2.1} & 48.5{\scriptsize$\pm$2.8}\gain{+6.4} & \textbf{58.3}{\scriptsize$\pm$1.8}\gain{+16.2} & 28.9{\scriptsize$\pm$1.5} & 29.5{\scriptsize$\pm$1.5}\gain{+0.6} & \textbf{36.9}{\scriptsize$\pm$1.4}\gain{+8.1} \\
\quad Gamma               & 40.4{\scriptsize$\pm$1.4} & 51.5{\scriptsize$\pm$1.3}\gain{+11.1} & \textbf{62.4}{\scriptsize$\pm$1.6}\gain{+22.0} & 28.1{\scriptsize$\pm$0.8} & 30.9{\scriptsize$\pm$0.7}\gain{+2.8} & \textbf{37.9}{\scriptsize$\pm$1.1}\gain{+9.7} \\
\quad Hue                 & 35.0{\scriptsize$\pm$1.1} & 46.0{\scriptsize$\pm$2.9}\gain{+11.0} & \textbf{50.1}{\scriptsize$\pm$2.5}\gain{+15.1} & 24.7{\scriptsize$\pm$0.8} & 29.0{\scriptsize$\pm$1.8}\gain{+4.3} & \textbf{32.6}{\scriptsize$\pm$1.7}\gain{+7.9} \\
\quad Contrast            & 48.6{\scriptsize$\pm$2.6} & 61.7{\scriptsize$\pm$3.0}\gain{+13.1} & \textbf{70.7}{\scriptsize$\pm$2.2}\gain{+22.1} & 33.8{\scriptsize$\pm$1.5} & 38.0{\scriptsize$\pm$1.9}\gain{+4.2} & \textbf{45.9}{\scriptsize$\pm$1.5}\gain{+12.1} \\
\quad Translation         & 40.2{\scriptsize$\pm$0.9} & 46.4{\scriptsize$\pm$2.0}\gain{+6.2} & \textbf{51.7}{\scriptsize$\pm$1.4}\gain{+11.5} & 28.3{\scriptsize$\pm$0.9} & 29.8{\scriptsize$\pm$1.5}\gain{+1.5} & \textbf{33.6}{\scriptsize$\pm$0.7}\gain{+5.3} \\
\bottomrule
\end{tabular}
\end{subtable}
\vspace{0.5em}
\begin{subtable}{\textwidth}
\centering
\caption{Predictor and encoder adapted.}
\begin{tabular}{l ccc @{\hspace{9pt}} ccc}
\toprule
 & \multicolumn{3}{c}{Final} & \multicolumn{3}{c}{Mean over replans} \\
\cmidrule(lr){2-4}\cmidrule(lr){5-7}
Setting & Frozen & AdaJEPA & OpTTA & Frozen & AdaJEPA & OpTTA \\
\midrule
In distribution           & 59.2{\scriptsize$\pm$2.1} & 74.0{\scriptsize$\pm$2.1}\gain{+14.8} & \textbf{83.2}{\scriptsize$\pm$2.4}\gain{+24.0} & 42.9{\scriptsize$\pm$1.5} & 50.9{\scriptsize$\pm$1.9}\gain{+8.0} & \textbf{56.4}{\scriptsize$\pm$1.7}\gain{+13.5} \\
\midrule
\multicolumn{7}{l}{\textit{Visual corruptions}} \\
\quad Blur                & 42.5{\scriptsize$\pm$0.8} & 64.9{\scriptsize$\pm$2.1}\gain{+22.4} & \textbf{71.6}{\scriptsize$\pm$1.2}\gain{+29.1} & 30.0{\scriptsize$\pm$0.8} & 40.1{\scriptsize$\pm$1.1}\gain{+10.1} & \textbf{46.0}{\scriptsize$\pm$0.9}\gain{+16.0} \\
\quad Salt \& pepper      & 48.4{\scriptsize$\pm$2.4} & 68.1{\scriptsize$\pm$2.7}\gain{+19.7} & \textbf{76.1}{\scriptsize$\pm$2.0}\gain{+27.7} & 32.4{\scriptsize$\pm$1.4} & 40.6{\scriptsize$\pm$1.4}\gain{+8.2} & \textbf{49.0}{\scriptsize$\pm$1.5}\gain{+16.6} \\
\quad Darkening           & 42.1{\scriptsize$\pm$2.1} & 48.3{\scriptsize$\pm$3.4}\gain{+6.2} & \textbf{58.8}{\scriptsize$\pm$2.3}\gain{+16.7} & 28.9{\scriptsize$\pm$1.5} & 29.5{\scriptsize$\pm$2.0}\gain{+0.6} & \textbf{36.9}{\scriptsize$\pm$1.7}\gain{+8.1} \\
\quad Gamma               & 40.4{\scriptsize$\pm$1.4} & 54.7{\scriptsize$\pm$0.8}\gain{+14.3} & \textbf{65.9}{\scriptsize$\pm$0.9}\gain{+25.5} & 28.1{\scriptsize$\pm$0.8} & 32.9{\scriptsize$\pm$0.5}\gain{+4.7} & \textbf{40.6}{\scriptsize$\pm$0.6}\gain{+12.5} \\
\quad Hue                 & 35.0{\scriptsize$\pm$1.1} & 45.7{\scriptsize$\pm$2.4}\gain{+10.7} & \textbf{50.8}{\scriptsize$\pm$3.0}\gain{+15.8} & 24.7{\scriptsize$\pm$0.8} & 28.9{\scriptsize$\pm$1.7}\gain{+4.2} & \textbf{32.3}{\scriptsize$\pm$1.8}\gain{+7.6} \\
\quad Contrast            & 48.6{\scriptsize$\pm$2.6} & 61.9{\scriptsize$\pm$2.8}\gain{+13.3} & \textbf{72.3}{\scriptsize$\pm$2.0}\gain{+23.7} & 33.8{\scriptsize$\pm$1.5} & 38.1{\scriptsize$\pm$1.5}\gain{+4.3} & \textbf{46.2}{\scriptsize$\pm$1.6}\gain{+12.4} \\
\quad Translation         & 40.2{\scriptsize$\pm$0.9} & 50.4{\scriptsize$\pm$2.2}\gain{+10.2} & \textbf{55.4}{\scriptsize$\pm$1.5}\gain{+15.2} & 28.3{\scriptsize$\pm$0.9} & 31.7{\scriptsize$\pm$1.7}\gain{+3.4} & \textbf{35.3}{\scriptsize$\pm$1.4}\gain{+7.0} \\
\bottomrule
\end{tabular}
\end{subtable}
\end{table}

\FloatBarrier

\end{document}